\documentclass[10pt,twocolumn,letterpaper]{article}

\usepackage{makecell} 
\usepackage{wrapfig}
\usepackage{caption}
\usepackage{cuted}
\usepackage{float}
\usepackage[table]{xcolor}  
\definecolor{rowgray}{gray}{0.92}
\definecolor{middlegray}{gray}{0.7}
\usepackage{multirow}

\usepackage{multirow, booktabs, xcolor, graphicx}

\usepackage{cvpr}              %

\usepackage{microtype}
\usepackage{pifont}
\newcommand{\Letter}{\ding{41}}

\definecolor{cvprblue}{rgb}{0.21,0.49,0.74}
\usepackage[pagebackref,breaklinks,colorlinks,allcolors=cvprblue]{hyperref}

\def\paperID{1700} %
\def\confName{CVPR}
\def\confYear{2026}

\title{VideoCanvas: Unified Video Completion from Arbitrary Spatiotemporal Patches via In-Context Conditioning}
\author{
Minghong Cai$^{1\dagger}$ \quad
Qiulin Wang$^{2\text{\,\Letter}}$ \quad
Zongli Ye$^{1}$ \quad
Wenze Liu$^{1}$ \quad
Quande Liu$^{2}$ \quad
Weicai Ye$^{2}$ \\
Xintao Wang$^{2}$ \quad
Pengfei Wan$^{2}$ \quad
Kun Gai$^{2}$ \quad
Xiangyu Yue$^{1\text{\,\Letter}}$ \\
$^{1}$MMLab, The Chinese University of Hong Kong \quad
$^{2}$Kling Team, Kuaishou Technology
}

\begin{document}

\twocolumn[{%
\maketitle
\vspace{-27pt}
\begin{figure}[H]
\hsize=\textwidth %
\centering
\includegraphics[width=0.98\textwidth]{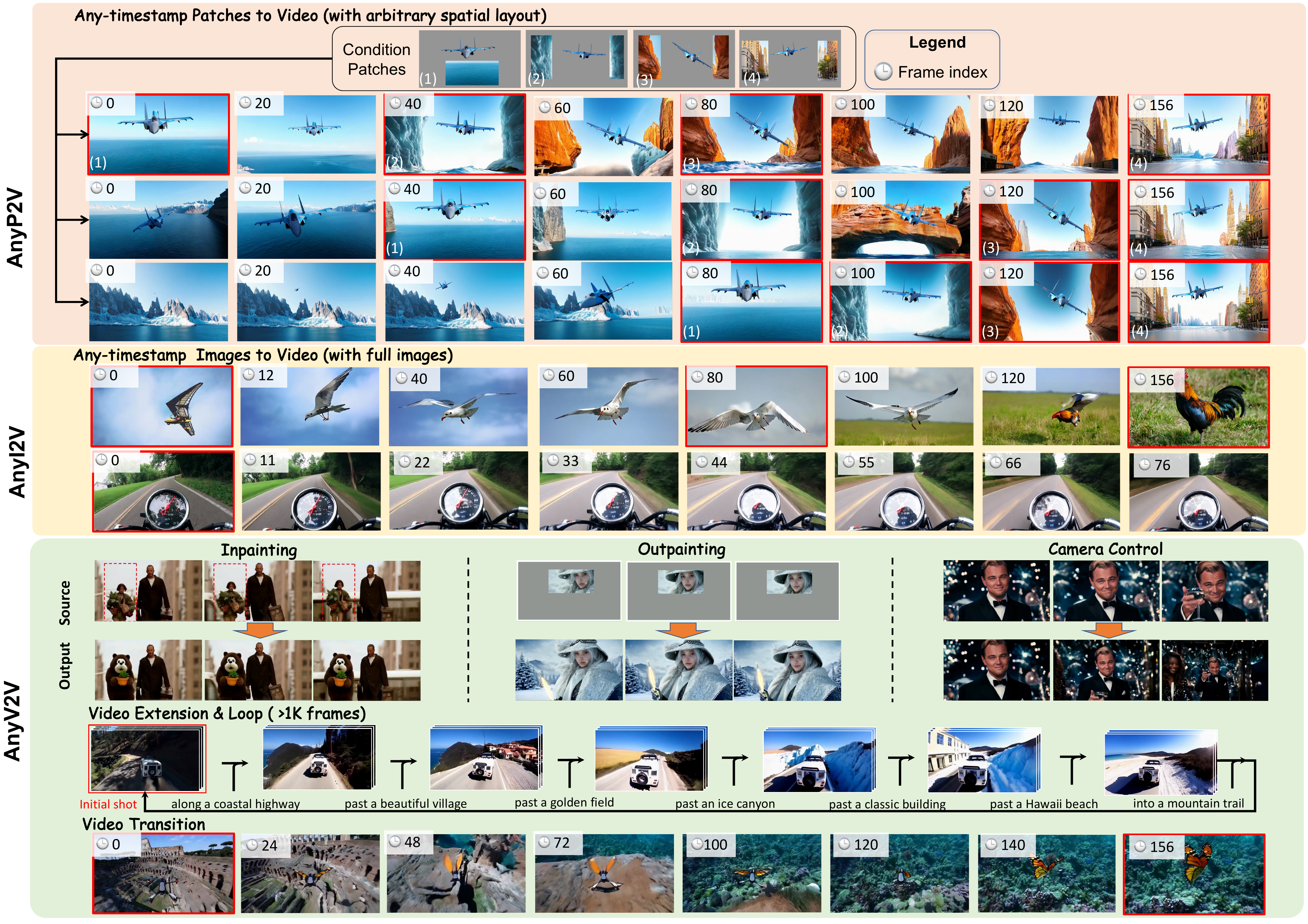}
\vspace{-1em}

\caption{
\textbf{VideoCanvas: Arbitrary Spatio-Temporal Video Completion.}
Given any conditions (frames or patches, outlined in \textcolor{red}{red}), the model fills in the remaining \textcolor{middlegray}{gray} regions to generate coherent, high-quality videos.
This unified formulation subsumes various tasks such as Any-Timestep-Patch/Image-to-Video, In/Outpainting, Camera Control, and Cross-scene Video Transitions, all in a zero-shot manner.
More results are available on our \textbf{Project Page}: \url{https://onevfall.github.io/project_page/videocanvas/}.
\textbf{Best viewed zoomed in.}
}
\label{fig: teaser}
\end{figure}
}]
\renewcommand{\thefootnote}{}
\footnotetext{\hspace{-1.8em}$^\dagger$Work done at Kuaishou Technology. \Letter\,Corresponding authors.}
\renewcommand{\thefootnote}{\arabic{footnote}}

\begin{abstract}
Existing controllable video generation methods are typically designed for rigid, task-specific settings---such as first-frame image-to-video, inpainting, or interpolation---treating spatio-temporal control as a set of isolated problems.
We formalize a unified task, \textit{arbitrary spatio-temporal video completion}, where a model generates a coherent video from user-specified patches placed at any spatial location and timestamp.
However, realizing such a unified framework within modern latent video diffusion models is non-trivial: causal video VAEs compress multiple frames into a single latent slot, making frame-level conditioning fundamentally ill-posed, and directly feeding sparsely populated, zero-padded video inputs into the VAE leads to severe out-of-distribution artifacts.
To address these challenges, we propose \textbf{VideoCanvas}, a simple yet effective framework that adapts the In-Context Conditioning paradigm to arbitrary spatio-temporal completion without modifying or retraining the VAE.
Our key idea is a hybrid conditioning strategy that decouples spatial and temporal control: spatially, we encode zero-padded full-frame canvases in image mode to keep VAE inputs in-distribution, and temporally we use Temporal RoPE Interpolation to assign each condition a continuous fractional index in the latent sequence for precise frame-level alignment.
To evaluate this capability, we develop VideoCanvasBench, the first benchmark for arbitrary spatio-temporal video completion, covering both intra-scene fidelity and inter-scene creativity.
Extensive experiments demonstrate that VideoCanvas achieves state-of-the-art performance across a diverse range of video generation tasks under a single, unified framework.
\end{abstract}

\section{Introduction}

Video generation has made significant strides with the advent of Diffusion Transformers (DiTs)~\citep{dit, chen2023videocrafter1, wan2025, yang2024cogvideox}, marking a turning point in the field's ability to synthesize high-quality videos.
However, generating videos that truly align with user intent remains a significant challenge.
Existing controllable approaches are typically constrained by rigid, task-specific formats—for example, conditioning only on a first frame~\citep{guo2023animatediff, kong2024hunyuanvideo}, using an initial clip with limited temporal horizon~\citep{bar2025navigation, yang2025resim}, or performing structural inpainting and outpainting~\citep{zhou2023propainter, wang2024your, yang2025gencompositor}.
These methods treat spatio-temporal control as a set of isolated problems, lacking a unified approach.
We propose a unified approach to bridge these fragmented tasks: \textit{treating video synthesis as painting on a spatio-temporal canvas}.
In this framework, users can place arbitrary content patches at any location and timestamp, and the model will synthesize a complete, temporally consistent video around them, as illustrated in Fig.~\ref{fig: teaser}.
This fine-grained control enables a wide range of applications, from creative content generation to practical use cases, such as reconstructing videos from partially transmitted or corrupted data packets~\citep{li2023reparo, du2020server}, or generating videos with specific spatial and temporal conditions for diverse domains.

\begin{figure*}[t]
  \centering
  \includegraphics[width=2\columnwidth]{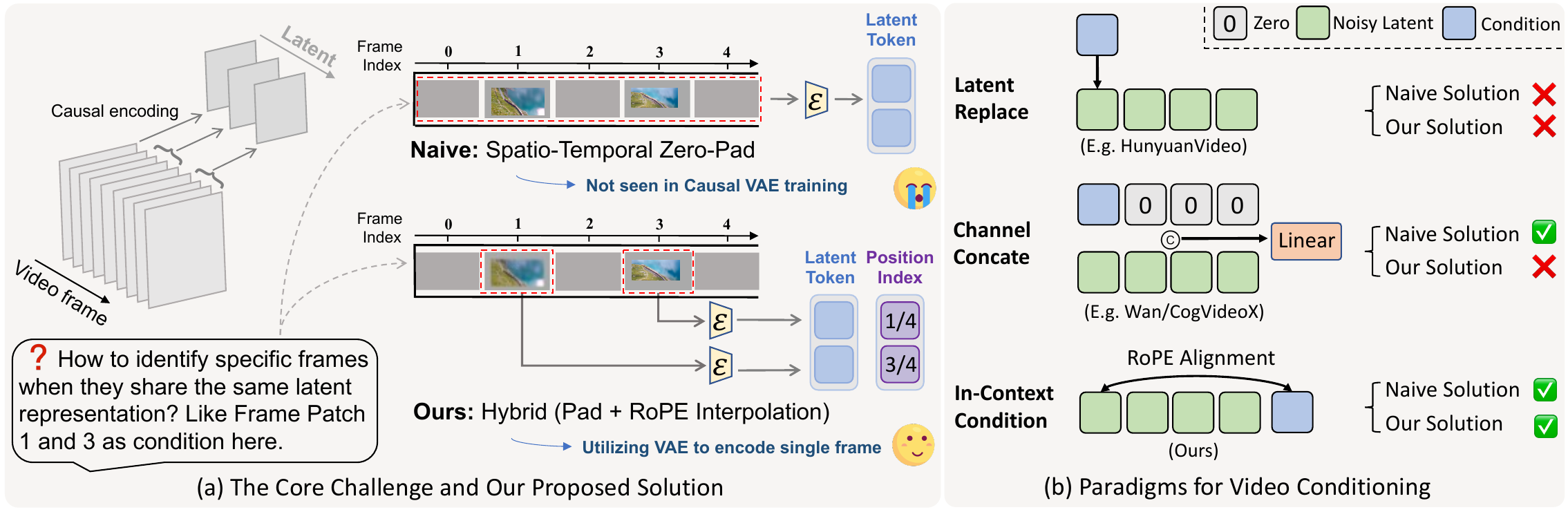}
\caption{
  \textbf{Core challenge and solution for pixel-frame-aware conditioning.}
  \textbf{(a)} Causal VAEs create temporal ambiguity by mapping frames to a single latent. We propose a hybrid solution combining Spatial Padding with Temporal RoPE Interpolation.
  \textbf{(b)} We show how competing paradigms are ill-suited for fine-grained control, while our \textit{ICC} approach provides an effective solution.
}

\vspace{-4mm}
  \label{fig:intro_figure}
\end{figure*}

Realizing this vision presents fundamental challenges across both spatial and temporal dimensions. Temporally, causal video VAEs compress multiple pixel frames into a single latent slot, creating indexing ambiguity—precisely the source of the difficulty in achieving frame-accurate control—as illustrated in Fig.~\ref{fig:intro_figure}(a).
Spatially, conditions may take arbitrary forms—from full frames to small, irregular patches—requiring a mechanism that can seamlessly unify inpainting and outpainting within one formulation.  
The core difficulty lies in designing a conditioning paradigm that can resolve both temporal ambiguity and spatial irregularity simultaneously.

Viewed through this lens, the limitations of existing paradigms become clear.  
Latent Replacement~\citep{hacohen2024ltx, kong2024hunyuanvideo} was designed mainly for first-frame I2V but fails to generalize, as it overwrites entire latent slots and disrupts temporal consistency once applied to arbitrary timestamps.  
Channel Concatenation and Adapter-style injection methods~\citep{yang2024cogvideox, wan2025, mou2024t2i, zhang2023adding} fuse conditional features either by concatenating at the input or injecting via lightweight encoders.  
Despite architectural differences, these approaches remain coarse-grained: pixel-frame-aware control ultimately requires feeding zero-padded frames to the VAE, but pretrained VAEs are not robust to such inputs.  
Making them work would require expensive VAE fine-tuning and re-training of the DiT backbone.  
More recent In-Context Conditioning (ICC) methods~\citep{tan2024ominicontrol, ju2025fulldit, he2025fulldit2, ye2025unic,guo2025long} inherit the same difficulty when naively combined with zero-padding: they still demand VAE/DiT re-training to handle the distribution shift, and further double the sequence length by encoding padded frames, resulting in severe inefficiency during both training and inference.

In this paper, we introduce \textbf{VideoCanvas}, the first framework to apply \emph{In-Context Conditioning} to the challenging task of arbitrary spatio-temporal video completion. We also propose a hybrid conditioning strategy that decouples space and time: spatial alignment is achieved by zero-padded VAE encoding of arbitrary patches, while temporal ambiguity is resolved by our novel RoPE Interpolation, which assigns continuous fractional indices to conditional frame tokens.  
This design removes the need for costly re-training of the VAE or architectural modifications of the DiT backbone, while allowing efficient fine-tuning to enable fine-grained pixel-frame-aware control within a simple, parameter-free ICC architecture.

To evaluate this unified task and framework, we present \textbf{VideoCanvasBench}, a comprehensive benchmark tailored for arbitrary spatio-temporal video completion.  
To the best of our knowledge, it is the first
to systematically incorporate multi-frame, non-homologous image and patch conditions to test both
intra-scene fidelity and inter-scene creativity. Our contributions are as follows:

\begin{itemize}[left=0pt]
    \item We formalize the task of \textit{arbitrary spatio-temporal video completion}, under which existing tasks such as image-to-video, interpolation, and inpainting emerge as special cases, and provide an in-depth analysis revealing the structural limitations of existing conditioning paradigms under this setting.
    \item We propose VideoCanvas, the first framework to adapt In-Context Conditioning to this task, with a hybrid strategy: Spatial Zero-Padding for in-distribution VAE encoding and Temporal RoPE Interpolation for precise frame-level alignment---enabling pixel-frame-aware control on frozen VAEs without retraining or new parameters.
    \item We introduce VideoCanvasBench, the first benchmark for arbitrary spatio-temporal completion, and demonstrate state-of-the-art performance across diverse settings.  
\end{itemize}

\section{Methodology}
\label{sec:method}

\subsection{Task Definition}

To enable flexible and unified video generation, we formalize the task of \textbf{arbitrary spatio-temporal video completion}.
Let a video be denoted as $\boldsymbol{X} = \{\boldsymbol{x}_0, \boldsymbol{x}_1, \dots, \boldsymbol{x}_{T-1}\}$ with $T$ frames.
A user provides a set of spatio-temporal conditions $\mathcal{P} = \{(\boldsymbol{p}_i, \boldsymbol{m}_i, t_i)\}_{i=1}^M$, where $\boldsymbol{p}_i$ is an image, $\boldsymbol{m}_i$ is a spatial mask specifying its placement within a frame, $t_i \in [0, T{-}1]$ is the temporal index, and $M$ is the number of conditions.
The goal is to generate a coherent video $\hat{\boldsymbol{X}}$ such that
\[
\hat{\boldsymbol{X}}[t_i] \odot \boldsymbol{m}_i \;\approx\; \boldsymbol{p}_i, \quad \forall i \in \{1, \dots, M\},
\]
while completing all unconditioned regions with plausible content.
This formulation naturally unifies many prior settings as special cases:
\textit{image-to-video} ($\mathcal{P}$ contains one full frame),
\textit{interpolation} ($\mathcal{P}$ specifies first and last frames),
and \textit{inpainting/outpainting} ($\mathcal{P}$ contains masked regions).
By allowing arbitrary spatial masks at arbitrary timestamps, it goes strictly beyond these rigid formats.

\subsection{Preliminaries}
\label{sec:preliminaries}

\noindent \textbf{Video DiT with 3D RoPE.}
Our work builds upon a latent video diffusion model that uses a Diffusion Transformer (DiT) backbone~\citep{dit} and is trained with a flow matching objective~\citep{lipman2022flow}.
To handle the spatio-temporal nature of video data, the model's self-attention mechanism is equipped with 3D Rotary Positional Embeddings (RoPE)~\citep{su2024roformer}.
The 3D RoPE encodes each token's position by applying a rotation to the query and key vectors in the attention mechanism, driven by the token's coordinates in 3D space (time, height, and width), allowing the model to capture both temporal continuity and spatial relationships. This mechanism is central to our ability to perform precise temporal alignment in video generation, as RoPE naturally supports interpolation to non-integer positions.

\noindent \textbf{Hybrid Video VAE.}
Modern video foundation models employ a Hybrid Video VAE~\citep{zhao2024cv,yang2024cogvideox,wu2025improved} that supports both image and video modes.
In video mode, the encoder performs causal temporal compression with a fixed stride $N$ (e.g., $N=4$): the first frame maps to latent index 0, and every subsequent $N$ frames collapse into a single latent slot.
Formally, for a pixel-frame index $i$, the latent index is $\lceil i/N \rceil$.
This stride-based compression is efficient but introduces \textit{pixel-frame ambiguity}: multiple frames (e.g., frames 1, 2, 3) may share the same latent representation, making precise frame-level conditioning non-trivial.

\begin{figure}[b]
  \vspace{-5mm}
  \centering
  \includegraphics[width=0.95\linewidth]{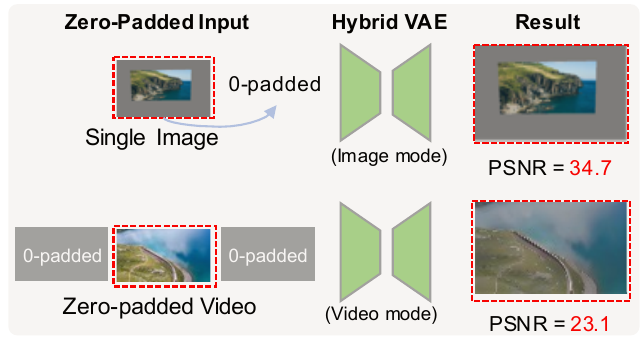}
  \vspace{-2mm}
  \caption{
      \textbf{Analysis of Zero-Padding VAE Reconstruction.}
    }
  \label{fig:padding}
  \vspace{-3mm}
\end{figure}

\subsection{Zero Padding Observations and Analysis}
\label{method:zero_pad_analysis}
Existing conditioning approaches~\citep{ju2024brushnet, bian2025videopainter, jiang2025vace} commonly represent missing content by zero-filling unobserved regions or frames. However, the impact of such zero-padded inputs on the causal VAEs of modern video models has not been thoroughly investigated. We analyze this in Fig.~\ref{fig:padding} (more details in Appendix~\ref{app:zero_padding}) and present two key findings:

\noindent \textbf{Spatial padding is robust in image mode.}
Image-based pipelines (e.g., BrushNet~\citep{ju2024brushnet} and VideoPainter~\citep{bian2025videopainter}) routinely process frames with large masked/blank regions, and image VAEs are typically robust to such spatial sparsity.
In our setting, we place patches on a zero-padded full-frame canvas and encode each frame with the video VAE in \emph{image} mode.
As shown in Fig.~\ref{fig:padding} (top), reconstruction fidelity drops only slightly, indicating that spatial zero-padding does not substantially shift the input distribution for the encoder.

\noindent \textbf{Temporal padding is harmful in video mode.}
Causal video VAEs compress $N$ consecutive frames into one latent slot (e.g., $N=4$).
If a conditioning sequence is formed by inserting zero-filled frames, then blank and non-blank frames are mixed within the same $N$-frame window, producing temporal dynamics that the VAE never observes during training.
Consequently, temporal zero-padding causes severe out-of-distribution degradation: Fig.~\ref{fig:padding} (bottom) shows large PSNR drops and visible corruption, consistently across strong backbones such as HunyuanVideo~\citep{kong2024hunyuanvideo} and CogVideoX~\citep{yang2024cogvideox}.

\begin{figure*}[t!]
  \centering
  \includegraphics[width=0.95\textwidth]{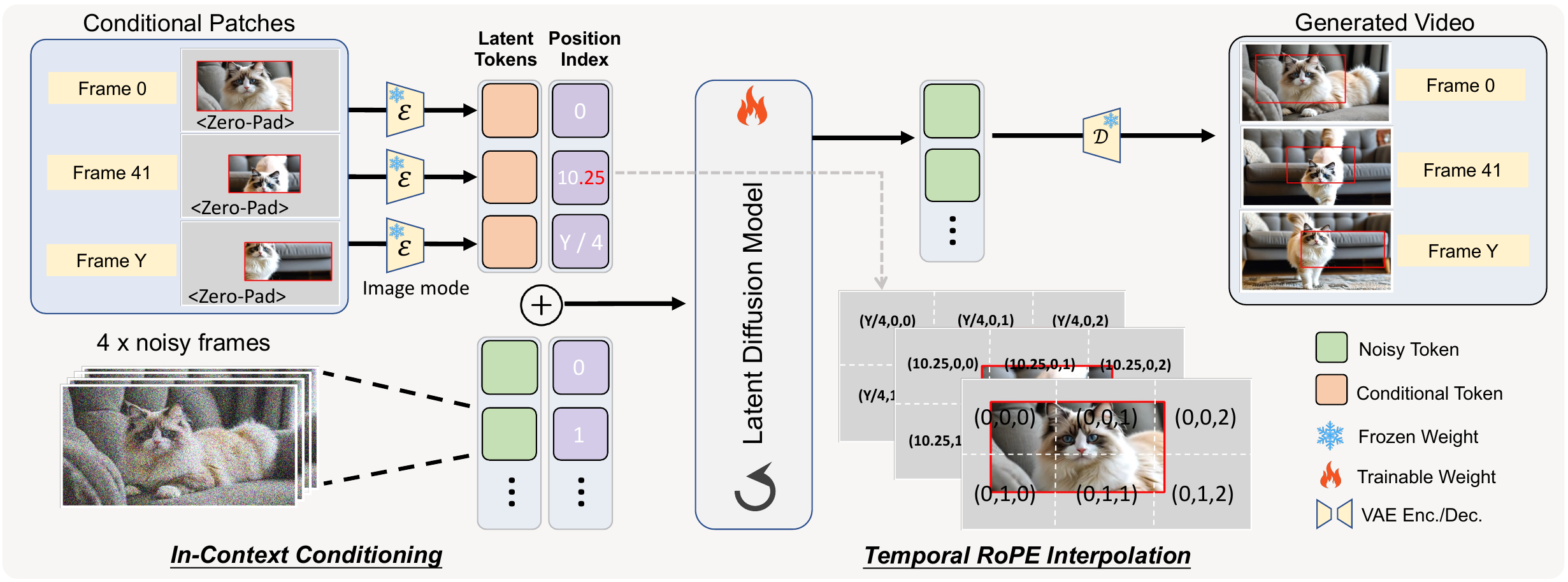}
    \caption{
        \textbf{The VideoCanvas pipeline.}
        We fine-tune a base T2V model to enable arbitrary spatio-temporal control, without introducing any additional parameters.
        Our framework leverages the In-Context Conditioning (ICC) paradigm. After preparing conditional patches with zero-padding for spatial placement, we use independent VAE encoding for temporal decoupling. Our RoPE Interpolation then aligns each discrete token by mapping its source pixel-frame index $Y$ to a fractional position $Y/N$, where $N$ is the VAE temporal stride (here, $N=4$). As illustrated, this maps Frame $41$ to position $10.25$. This strategy enables fine-grained control without architectural changes.
    }
    \vspace{-3mm}
  \label{fig:main_pipeline}
\end{figure*}

\subsection{Architecture Design Space Analysis}
\label{sec:arch_analysis}
Given that spatial padding is VAE-friendly while temporal padding is destructive (Sec.~\ref{method:zero_pad_analysis}), a natural question arises: which conditioning paradigm can best exploit this asymmetry? We revisit three common approaches under two solution strategies, as illustrated in Fig.~\ref{fig:intro_figure}(b).

The \textbf{naive solution} constructs a full-length video with zero-filled unobserved frames and processes it through the VAE in video mode.
\textit{Latent Replacement}~\citep{kong2024hunyuanvideo} directly overwrites latent slots, but since it operates after VAE encoding, it cannot control which pixel frames contribute to each slot---making arbitrary-timestamp conditioning infeasible even under this strategy.
\textit{Channel Concatenation}~\citep{yang2024cogvideox, wan2025} can adopt this approach by concatenating the zero-padded conditional sequence along the channel dimension, but doing so inevitably triggers the temporal zero-padding degradation identified in Sec.~\ref{method:zero_pad_analysis}.
\textit{ICC} also supports this strategy, but inherits the same VAE quality loss.

\textbf{Our solution} instead encodes each conditional frame independently in image mode, bypassing temporal zero-padding entirely.
This requires the paradigm to accept standalone tokens with flexible temporal positions.
Latent Replacement cannot accommodate independently encoded tokens, as it relies on overwriting fixed slots in the noisy latent tensor.
Channel Concatenation is similarly constrained to the integer latent grid, offering no mechanism for fractional temporal alignment.
Only ICC, which concatenates conditions along the token dimension, naturally supports both independent encoding and continuous-time positional assignments via RoPE interpolation.
We therefore build VideoCanvas upon the In-Context Conditioning paradigm, as detailed in Sec.~\ref{sec:videocanvas_method}.

\subsection{VideoCanvas Pipeline}
\label{sec:videocanvas_method}

To address the challenge of arbitrary spatio-temporal completion, we propose VideoCanvas, a unified framework built upon the In-Context Conditioning (ICC) paradigm.
As established in Sec.~\ref{sec:arch_analysis}, ICC uniquely enables sub-latent temporal precision through flexible positional assignments.
Combined with the VAE's tolerance for spatial padding (Sec.~\ref{method:zero_pad_analysis}), we introduce a hybrid conditioning strategy that decouples spatial and temporal alignment, enabling fine-grained, pixel-frame-aware control on a frozen VAE and a fine-tuned DiT with zero new parameters.
The entire pipeline is illustrated in Fig.~\ref{fig:main_pipeline}.

\paragraph{Spatial Conditioning via Zero-Padding.}
As shown on the left of Fig.~\ref{fig:main_pipeline}, our process begins at the pixel level.
For each conditional patch $(\boldsymbol{p}_i, \boldsymbol{m}_i, t_i) \in \mathcal{P}$, we construct a full-frame canvas, place the patch $\boldsymbol{p}_i$ in its correct spatial location according to mask $\boldsymbol{m}_i$, and fill the remaining pixels with zeros.
Formally, the prepared frame is $\boldsymbol{x}_{\text{prep}, i} = \boldsymbol{m}_i \odot \boldsymbol{p}_i$, where $\odot$ denotes element-wise multiplication and unconditioned regions are zero-filled.
This preserves the absolute positional information required for spatial control, and crucially, does not cause out-of-distribution issues as the VAE tolerates spatial padding well (Sec.~\ref{method:zero_pad_analysis}).

\paragraph{Temporal Decoupling via Independent VAE Encoding.}
Next, each of these prepared frames is encoded \textit{independently} by the frozen VAE in its image mode.
This is a critical step for temporal decoupling: by encoding each frame individually, we bypass the VAE's causal temporal compression mechanism.
The result is a set of conditional latent tokens $\boldsymbol{z}_{\text{cond}, i} = \mathcal{E}(\boldsymbol{x}_{\text{prep}, i})$, where each token purely represents its corresponding single pixel frame, free from the temporal ambiguity discussed in Sec.~\ref{sec:preliminaries}.

\paragraph{Temporal Alignment via RoPE Interpolation.}
\label{sec:rope_alignment}
Having obtained temporally decoupled conditional latents, we now address the core challenge: precisely aligning them within the DiT's 3D spatio-temporal grid.
Following the ICC paradigm, we construct a unified sequence by concatenating the conditional tokens with the target video latent.
Let $\boldsymbol{z}_{\text{source}}$ denote the latent representation of the ground-truth target video, obtained by encoding the full video with the VAE in video mode.
The complete input sequence is then:
\begin{equation*}
\boldsymbol{z} = \text{Concat}(\{\boldsymbol{z}_{\text{cond}, i}\}_{i=1}^M, \boldsymbol{z}_{\text{source}}).
\end{equation*}

We leverage the continuous nature of the 3D RoPE used by our DiT backbone.
In the standard implementation, for a token at integer latent index $k$, RoPE applies a rotation to the query vector $\boldsymbol{q} \in \mathbb{R}^{D_t}$.
For the $j$-th pair of elements, the transformation is:
\begin{equation}
\label{eq:std_rope}
\begin{pmatrix} q'_{2j} \\ q'_{2j+1} \end{pmatrix} =
\begin{pmatrix}
\cos(k \theta_j) & -\sin(k \theta_j) \\
\sin(k \theta_j) & \cos(k \theta_j)
\end{pmatrix}
\begin{pmatrix} q_{2j} \\ q_{2j+1} \end{pmatrix},
\end{equation}
where $\theta_j = b^{-2j/D_t}$ are the base frequencies.
This integer indexing $k \in \{0, 1, \dots\}$ suffices for the source latent $\boldsymbol{z}_{\text{source}}$, which aligns with the VAE's compressed grid.

However, our conditional tokens originate from specific pixel frames that may fall \textit{between} integer latent indices.
For example, with VAE stride $N=4$, a condition from Frame 41 should ideally be positioned between latent indices 10 and 11.
To achieve this sub-latent precision, we introduce \textbf{Temporal RoPE Interpolation}: given a condition from pixel-frame $\tau_i$, we assign it a fractional temporal position $k' = \tau_i / N$ and substitute into Eq.~\eqref{eq:std_rope}:
\begin{equation}
\label{eq:interp_rope}
\begin{pmatrix} \tilde{q}_{2j} \\ \tilde{q}_{2j+1} \end{pmatrix} =
\begin{pmatrix}
\cos(\frac{\tau_i}{N} \theta_j) & -\sin(\frac{\tau_i}{N} \theta_j) \\
\sin(\frac{\tau_i}{N} \theta_j) & \cos(\frac{\tau_i}{N} \theta_j)
\end{pmatrix}
\begin{pmatrix} q_{2j} \\ q_{2j+1} \end{pmatrix}.
\end{equation}
As illustrated in Fig.~\ref{fig:main_pipeline}, this maps Frame 41 to the fractional position $10.25$.
For $N=4$, pixel frames $\tau \in \{1, 2, 3\}$ within the first latent window map to fractional positions $\{0.25, 0.50, 0.75\}$, allowing the attention mechanism to perceive conditions at precise sub-latent timestamps.
This strategy enables pixel-frame-aware temporal control that is structurally inaccessible to other paradigms.

\paragraph{Training Objective.}
The DiT model, $\boldsymbol{f}_\theta$, is fine-tuned with this unified sequence under the flow matching objective~\citep{lipman2022flow,liu2022flow}. The noising process is applied to the source latent of the sequence. The model input at time $t$ is thus a combination between the clean condition and the noisy latent:
$\boldsymbol{z}_t = (1 - t) \boldsymbol{z}_{\text{source}} + t \boldsymbol{\epsilon}$. 
The model is trained to predict the velocity field, and the loss only supervises the non-conditional regions:
\begin{equation*}
\mathcal{L}_\text{FM}(\theta) = 
\mathbb{E}
\left[\left\Vert \boldsymbol{f}_\theta(\boldsymbol{z}_t, t, \boldsymbol{c}_{\text{text}}) - (-\boldsymbol{z}_{\text{source}} + \boldsymbol{\epsilon}) \big) \right\Vert^2 \right].
\end{equation*}
This objective trains the DiT to treat the conditional tokens as fixed context while generating a coherent completion for the target video.

\begin{table*}[t]
\centering
\scriptsize
\caption{Quantitative evaluations on our benchmark against state-of-the-art models. Our unified framework demonstrates strong and consistent performance across a wide range of tasks, excelling in challenging few-frame-to-video interpolation and maintaining competitiveness in specialized domains like inpainting.}
\label{tab:main_benchmark_results} %
\renewcommand{\arraystretch}{1.2}
\addtolength{\tabcolsep}{-1pt}
\begin{tabular}{p{0.75cm}|l|ccccccccccc}

\toprule
\multirow{3}{*}{Type} & \multirow{3}{*}{Method} & 
& \multicolumn{9}{c}{Video Quality \& Video Consistency} \\
\cmidrule(lr){4-12}
& & FVD$\downarrow$ &
Aesthetic & Background & Dynamic & Imaging & Motion & Overall &
Subject & Temporal & \textbf{Normalized} \\
& & &
Quality & Consistency & Degree & Quality & Smoothness & Consistency &
Consistency & Flickering & \textbf{Average} \\
\midrule

I2V
& CogVideoX-1.5~\citep{yang2024cogvideox}
& 14.980
& 57.10\% & 95.17\% & 21.00\%
& \textbf{74.30\%} & 99.03\% & \textbf{25.22}\%
& 95.32\% & \textbf{97.85\%} & 70.62\% \\

& HunyuanVideo~\citep{kong2024hunyuanvideo}
& 16.756
& \textbf{57.21\%} & 95.19\% & 27.00\%
& 73.31\% & \textbf{99.13\%}& 25.01\%
& 93.60\% & 97.47\% & 70.99\% \\

\rowcolor{rowgray}
\cellcolor{white} & \textbf{VideoCanvas} (Ours)
& \textbf{14.476}
& 56.05\% & \textbf{95.96\%} & \textbf{30.00\%}
& 72.54\% & 99.04\% & 25.21\%
& \textbf{95.74\%} & 96.98\% & \textbf{71.44\%} \\
\midrule

FLF2V
& CogVideoX-FT~\citep{CogVideoX-FT}
& 13.545
& 56.88\% & 95.04\% & 18.00\%
& 72.65\% & 98.89\% & 25.00\%
& 95.55\% & 96.89\% & 69.86\% \\

& Sci-Fi~\citep{chen2025sci}
& 13.181
& \textbf{57.00\%} & 95.28\% & 9.00\%
& \textbf{72.97\%} & \textbf{99.11\%} & 25.11\%
& \textbf{95.83\%} & \textbf{97.35\%} & 68.96\% \\

\rowcolor{rowgray}
\cellcolor{white} & \textbf{VideoCanvas} (Ours)
& \textbf{9.053}
& 55.91\% & \textbf{96.01\%} & \textbf{42.00\%}
& 72.51\% & 98.97\% & \textbf{25.15\%}
& 95.44\% & 96.87\% & \textbf{72.86\%} \\
\midrule

TF2V
& CogVideoX-FT~\citep{CogVideoX-FT}
& 7.522
& 55.24\% & 94.82\% & 35.00\%
& 71.84\% & 99.05\% & 24.82\%
& 95.66\% & 97.36\% & 71.72\% \\

& Sci-Fi~\citep{chen2025sci}
& 9.523
& \textbf{57.87\%} & 94.59\% & 23.00\%
& \textbf{72.37\%} & \textbf{99.15\%} & 24.98\%
& \textbf{95.85\%} & \textbf{97.72\%} & 70.69\% \\

\rowcolor{rowgray}
\cellcolor{white} & \textbf{VideoCanvas} (Ours)
& \textbf{7.393}
& 55.76\% & \textbf{96.13\%} & \textbf{45.00\%}
& 72.01\% & 98.97\% & \textbf{25.13\%}
& 95.69\% & 96.93\% & \textbf{73.20\%} \\
\midrule

FLP2V
& Flux+Wan~\citep{flux2024,wan2025}
& 18.689
& \textbf{57.24\%} & 93.78\% & \textbf{67.00\%}
& \textbf{73.69\%} & 96.84\% & \textbf{25.25\%}
& 91.50\% & 94.49\% & \textbf{74.97\%} \\
\rowcolor{rowgray}
\cellcolor{white} & \textbf{VideoCanvas} (Ours)
& \textbf{16.799}
& 55.81\% & \textbf{96.08\%} & 55.00\%
& 72.27\% & \textbf{99.07\%}& 24.85\%
& \textbf{95.91}\% & \textbf{97.14\%} & 74.52\% \\
\midrule

Inpaint
& ProPainter~\citep{Zhou_2023_ICCV}
& 4.236
& 52.97\% & \textbf{96.26\%} & \textbf{49.00\%}
& 70.22\% & 98.62\% & 24.81\%
& 95.39\% & 96.83\% & 73.01\% \\

& VACE~\citep{jiang2025vace}
& \textbf{2.218}
& \textbf{56.72\%} & 95.55\% & \textbf{49.00\%}
& \textbf{74.81\%} & 98.22\% & 20.41\%
& 95.74\% & 96.24\% & 73.34\% \\

\rowcolor{rowgray}
\cellcolor{white} & \textbf{VideoCanvas} (Ours)
& 5.521
& 54.82\% & 96.06\% & \textbf{49.00\%}
& 72.60\% & \textbf{98.87\%} & \textbf{25.22\%}
& \textbf{95.88\%} & \textbf{97.05\%} & \textbf{73.69\%} \\
\midrule

Outpaint
& M3DDM~\citep{fan2023hierarchical}
& 75.266
& 39.35\% & 95.62\% & 35.00\%
& 52.13\% & 98.85\% & 11.24\%
& 94.77\% & \textbf{98.10\%} & 65.63\% \\

& VACE~\citep{jiang2025vace}
& \textbf{5.129}
& \textbf{56.82\%} & 95.63\% & 47.00\%
& \textbf{75.03\%} & 98.29\% & \textbf{25.30\%}
& \textbf{95.82\%} & 96.43\% & 73.79\% \\

\rowcolor{rowgray}
\cellcolor{white} & \textbf{VideoCanvas} (Ours)
& 9.119
& 55.88\% & \textbf{95.93\%} & \textbf{49.00\%}
& 72.78\% & \textbf{98.92\%} & 25.19\%
& 95.81\% & 97.07\% & \textbf{73.82\%} \\
\bottomrule
\end{tabular}
\end{table*}

\section{VideoCanvasBench}
\label{sec:benchmark}

Existing benchmarks focus on rigid tasks such as I2V or outpainting, and cannot assess the flexible spatio-temporal control central to our formulation. 
We therefore introduce \textit{VideoCanvasBench}, the first benchmark systematically designed for arbitrary spatio-temporal video completion. 

The benchmark probes two complementary capabilities: high-fidelity completion within a single scene (homologous) and creative synthesis across different sources (non-homologous). 
It consists of three categories:
(1) \textbf{AnyP2V}, using partial patches at fixed anchor timestamps (Start, Middle, End). We construct all seven possible combinations—single-frame (S, M, E), two-frame (S+M, S+E, M+E), and three-frame (S+M+E)—to evaluate interpolation fidelity under varying temporal sparsity.  
(2) \textbf{AnyI2V}, using full-frame conditions at the same timestamps, designed to test the completion of full-frame content.  
(3) \textbf{AnyV2V}, covering video-level completion scenarios such as inpainting, outpainting, and transitions between non-homologous clips.
In total, VideoCanvasBench comprises over \textbf{2,000} test cases. 
Further construction details are provided in Appendix~\ref{supp:benchmark_details}.

\section{Experiments}

\subsection{Setup}
\noindent \textbf{Backbone.}
We fine-tune a latent video diffusion model with a causal VAE (see Appendix~\ref{app:impl_details} for training details). Inference uses $50$ DDIM steps with a CFG scale of $7.5$.

\noindent \textbf{Compared Methods.} We compare our approach with state-of-the-art video generation methods, where the tasks include i) image-to-video (I2V) generation, such as CogVideoX-1.5~\citep{yang2024cogvideox} and HunyuanVideo~\citep{kong2024hunyuanvideo}; ii) first-last-frame-to-video (FLF2V) generation, such as CogVideoX-FT~\citep{CogVideoX-FT} and Sci-Fi~\citep{chen2025sci}; iii) first-middle-last-frame-to-video (TF2V), where we adapt CogVideoX-FT~\citep{CogVideoX-FT} and Sci-Fi~\citep{chen2025sci} by decomposing the task into two FLF2V subproblems; iv) first-last-patch-to-video (FLP2V) generation, which uses FLUX-Inpainting model~\citep{flux2024} to generate two frames at first, and then use video interpolation model~\citep{wan2025}; v) video inpainting and outpainting tasks such as VACE~\citep{jiang2025vace}. We provide an evaluation of overall video performance, including video fidelity (FVD~\citep{Unterthiner2018TowardsAG}) and a comprehensive suite of metrics from VBench~\citep{huang2024vbench} for video quality and consistency.

\subsection{Main Results}
\label{sec:main_results}

\begin{figure*}[t]
  \centering
  \includegraphics[width=\linewidth]{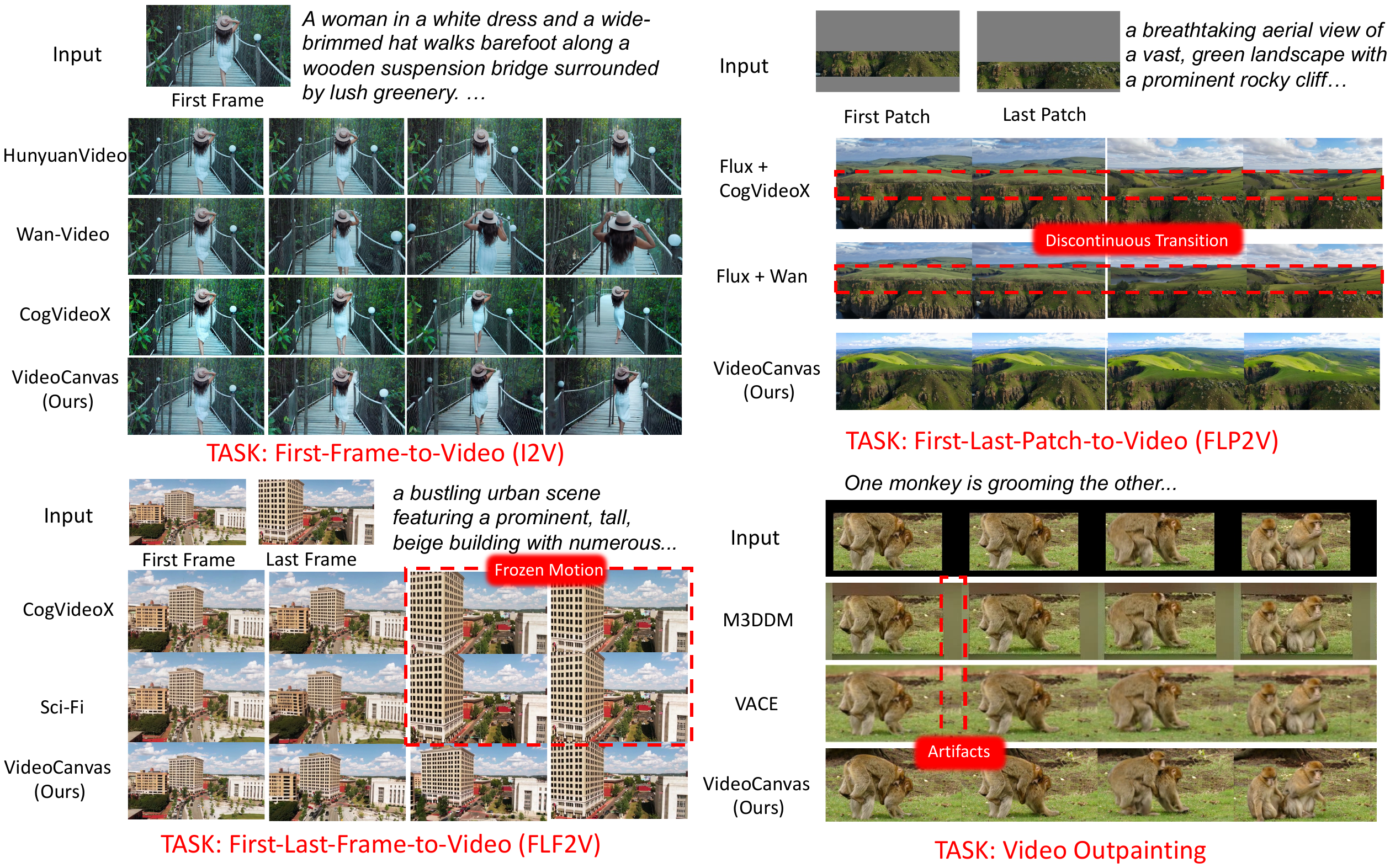}
  \vspace{-5mm}
  \caption{\textbf{Qualitative comparison with state-of-the-art methods} on four representative VideoCanvasBench tasks.}
  \label{fig:sota_comparison}
  \vspace{-3mm}
\end{figure*}

\begin{table}[t!]
\caption{Conditioning paradigm comparison. All methods are trained on the same backbone for fair comparison.}
\label{tab:paradigm_comparison}
\centering
\small
\setlength{\tabcolsep}{8pt}
\renewcommand{\arraystretch}{1.15}
\begin{tabular}{lccc}
\toprule
\textbf{Method} & FVD$\downarrow$ &
\makecell{Dynamic\\Degree$\uparrow$} & \makecell{Imaging\\Quality$\uparrow$}
\\
\midrule
Replacement        & 15.937 & 25.42 & 69.02 \\
Channel Concat     & 14.187 & 41.77 & 68.58 \\
\rowcolor{rowgray}
\textbf{ICC (Ours)} & \textbf{13.734} & \textbf{43.81} & \textbf{69.10} \\
\bottomrule
\end{tabular}
\vspace{-0.4cm}
\end{table}

\begin{figure}[t!]
  \centering
  \includegraphics[width=1\linewidth]{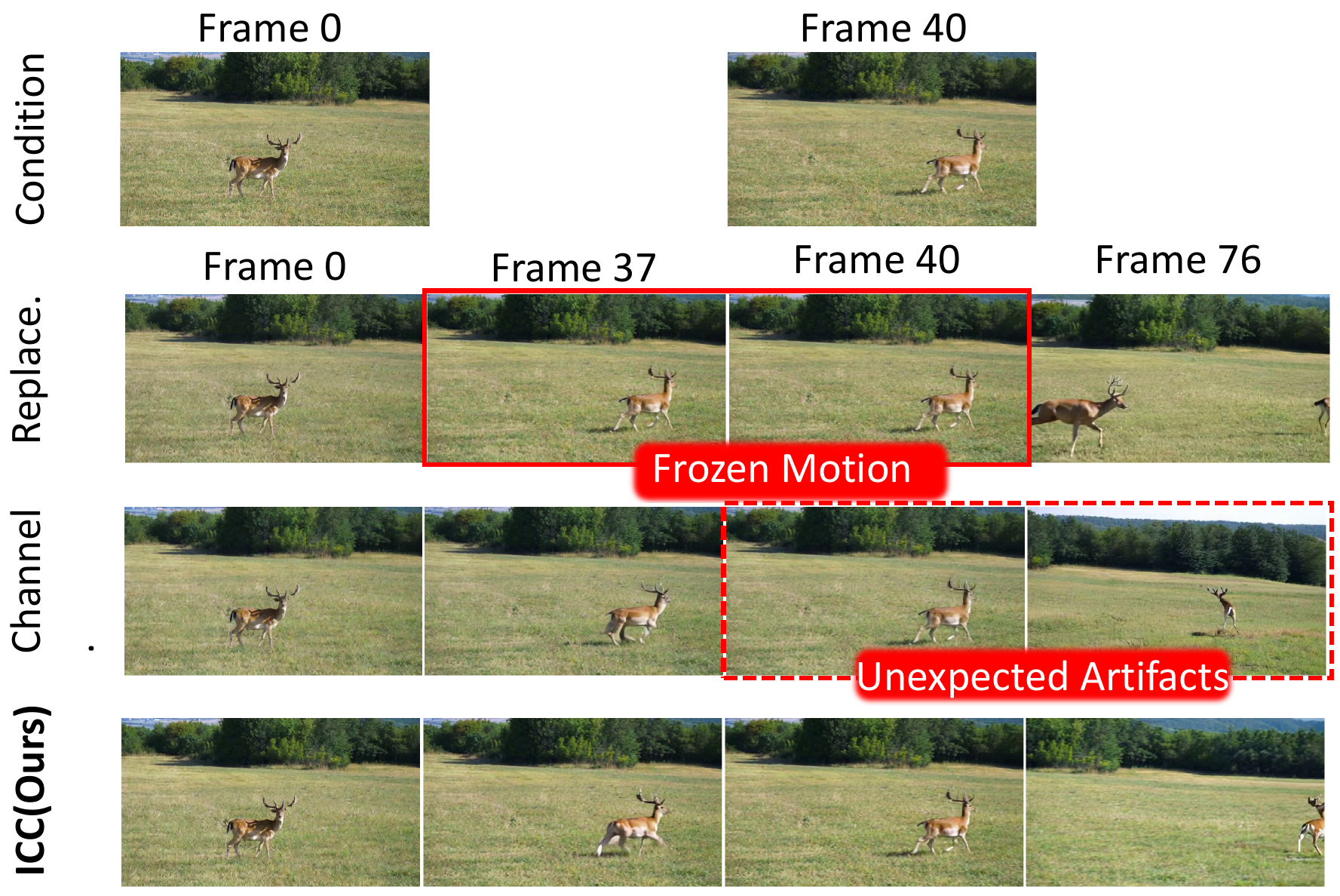}  
  \vspace{-5mm}
  \caption{\textbf{Qualitative comparison of conditioning paradigms} on a two-frame I2V task.}
  \label{fig:main_comparison}
  \vspace{-3mm}
\end{figure}

\noindent \textbf{SOTA Comparison on VideoCanvasBench.}
We evaluate our full model against leading specialized models on our benchmark. The results are summarized in Tab.~\ref{tab:main_benchmark_results}.
For standard \textbf{I2V}, our model achieves the lowest FVD and the highest consistency scores (Background and Subject Consistency), demonstrating strong distributional fidelity and identity preservation within a single unified framework.
The advantage of our approach becomes more pronounced in multi-frame interpolation tasks. In both \textbf{FLF2V} and \textbf{TF2V} settings, our method significantly outperforms competitors, achieving the lowest FVD and the highest Dynamic Degree. This highlights its superior capability in reasoning over sparse temporal conditions to generate coherent and dynamic motion.
In specialized tasks like \textbf{Inpainting} and \textbf{Outpainting}, dedicated models such as VACE show very strong performance, particularly in FVD. However, our unified framework remains highly competitive, often achieving the best normalized average score. This demonstrates the versatility of our approach, which can handle a wide range of completion tasks effectively without being specifically designed for any single one.

\noindent \textbf{Paradigm Comparison.}
We isolate the conditioning mechanism by comparing \emph{Latent Replacement}, \emph{Channel Concatenation}, and our \emph{In-Context Conditioning (ICC)} on the same backbone. As shown in Tab.~\ref{tab:paradigm_comparison}, ICC achieves the best (lowest) FVD, indicating the generated videos have the highest distributional similarity to real videos. Furthermore, it scores highest in both Imaging Quality and Dynamic Degree, demonstrating its superior ability to generate high-quality, dynamic content. Latent Replacement, while simple, struggles to synthesize motion, and Channel Concatenation lags behind ICC across all key metrics. This confirms that ICC is intrinsically a more effective paradigm for this task.

\begin{figure*}[t!]
  \centering
  \includegraphics[width=0.98\textwidth]{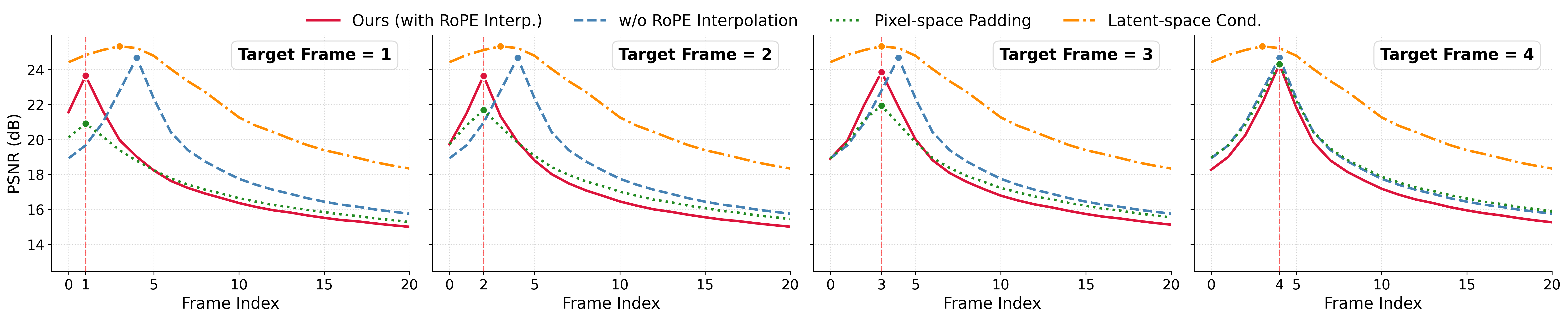}
  \vspace{-3mm}
  \caption{Ablation Study on Temporal Alignment Strategies.
    Per-frame PSNR for single-frame I2V with frame index targets $2$/$3$/$4$. 
    Our method (red) peaks exactly at the target frame. 
    ``w/o RoPE Interpolation'' (blue) misaligns, 
    ``Latent-space Condition'' (orange) collapses motion, 
    and ``Pixel-space Padding'' (green) is precise but degraded. Zoom in for best view.}
  \vspace{-3mm}
  \label{fig:psnr_spike}
\end{figure*}

\noindent \textbf{Qualitative Comparison.}
Fig.~\ref{fig:sota_comparison} provides per-task qualitative evidence.
For \textbf{FLF2V}, the test case involves high-dynamic, long-range interpolation between two distant keyframes; both CogVideoX-FT and Sci-Fi produce \textit{frozen motion} near the conditioning boundaries, failing to synthesize plausible intermediate dynamics, while our model generates smooth and coherent transitions.
For \textbf{FLP2V}, although Flux+Wan achieves competitive metrics, its two-stage pipeline (image generation followed by video interpolation) introduces \textit{discontinuous transitions} between the independently generated patches; our end-to-end framework produces seamless completions.
For \textbf{Outpainting}, M3DDM exhibits severe artifacts and VACE shows visible distortions in the expanded regions, whereas our model maintains high fidelity throughout.
We further compare conditioning paradigms in Fig.~\ref{fig:main_comparison}: Latent Replacement suffers from \textit{frozen motion}, while Channel Concatenation introduces \textit{unexpected artifacts}. Our ICC-based method generates smooth motion with preserved identity. More visualizations are in Appendix~\ref{app:qualitative}.

Overall, both quantitative and qualitative evidence show that our proposed framework, combining Temporal RoPE Interpolation with the In-Context Conditioning paradigm, establishes a new state of the art in the versatile and challenging task of arbitrary spatio-temporal video completion.

\subsection{Ablation Study}
\label{sec:ablation}

\begin{figure}[t]
  \centering
  \includegraphics[width=\linewidth]{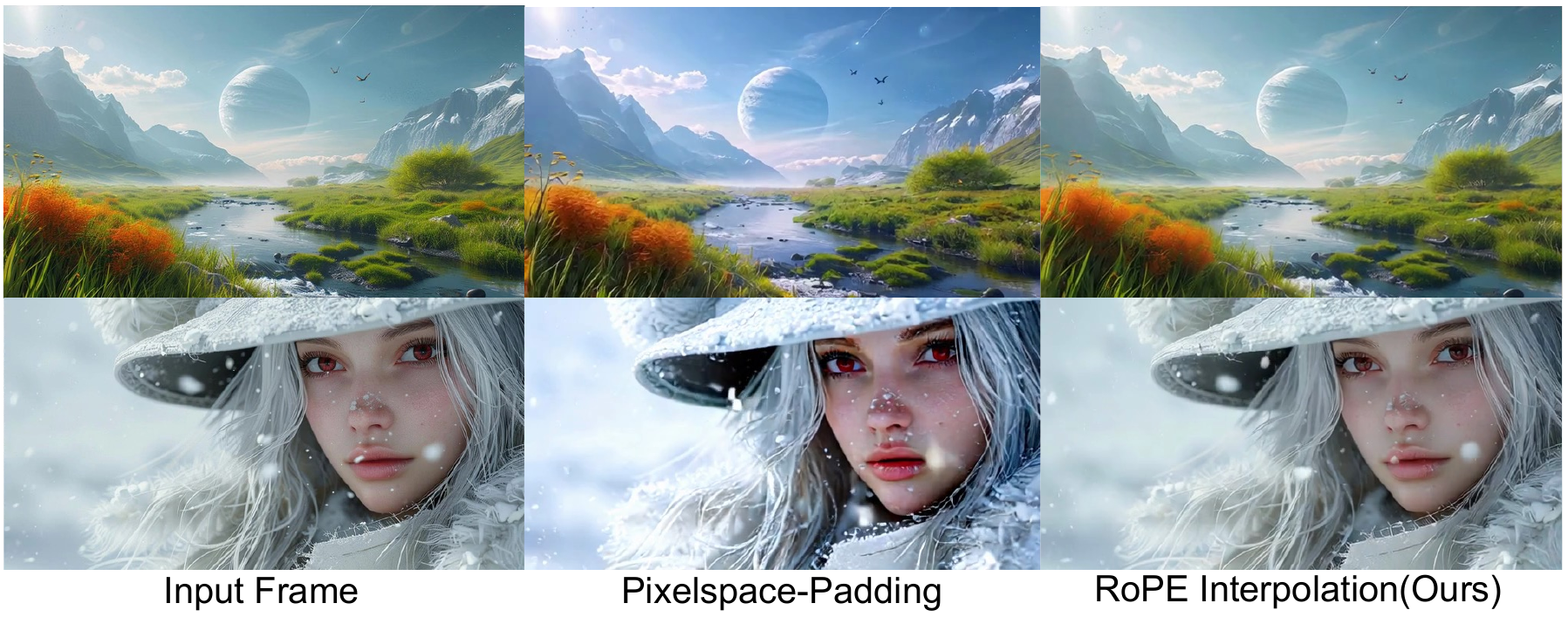}
  \vspace{-6mm}
  \caption{Pixelspace-Padding vs.\ RoPE Interpolation.}
  \label{fig:padding_vs_rope}
  \vspace{-2mm}
\end{figure}

As discussed in Fig.~\ref{fig:intro_figure} (a), causal video VAEs map several pixel frames into one latent, creating ambiguity when conditioning on a specific frame. To address this, we compare four alignment strategies: 
(i) \emph{Latent-space Conditioning}: encode the entire video with the VAE to obtain a latent sequence, then use the corresponding latent slice as condition.
(ii) \emph{Pixel-space Padding}: construct a video where non-target frames are zeroed, then encode it with the VAE.
(iii) \emph{w/o RoPE Interpolation}: encode each conditional frame independently (image mode) and assign it to the nearest discrete temporal slot.
(iv) \emph{Our full method with Temporal RoPE Interpolation}.

\noindent \textbf{Qualitative evidence.}  
Fig.~\ref{fig:padding_vs_rope} illustrates that while pixel-space padding is temporally precise, it introduces visible artifacts like color shifts and texture degradation because the VAE was not trained on zero-filled inputs. In contrast, our RoPE-based alignment preserves the conditional frame with high fidelity.

\noindent \textbf{Quantitative analysis.}  
We further evaluate single-frame I2V generation targeting frame indices $2, 3, 4$. As shown in Fig.~\ref{fig:psnr_spike} and Tab.~\ref{tab:ablation}, \emph{Latent-space Conditioning} achieves high PSNR but collapses motion (very low Dynamic Degree). \emph{w/o RoPE Interpolation} recovers dynamics but its PSNR peaks are misaligned. \emph{Pixel-space Padding} peaks correctly but suffers from lower fidelity. Our method with RoPE Interpolation not only aligns perfectly with the target frames but also achieves the best balance of fidelity and motion generation. These results confirm that our proposed strategy uniquely provides both fine-grained control and high-quality generation.

\begin{table}[t]
  \centering
  \caption{Ablation Study on Temporal Alignment Strategies. 
  }
  \label{tab:ablation}
  \small
  {%
  \setlength{\tabcolsep}{4pt}\renewcommand{\arraystretch}{1.15}
  \begin{tabular}{lccc}
    \toprule
    Method & PSNR$\uparrow$ & \makecell{Dynamic\\Degree$\uparrow$} & \makecell{Imaging\\Quality$\uparrow$} \\
    \midrule
    Ours (RoPE Interp.)        & \underline{23.86} & \textbf{39.75} & \textbf{71.61} \\
    w/o RoPE Interp.           & 22.95             & 23.00          & 70.85 \\
    Pixel-space Pad.           & 22.02             & \underline{30.25} & \underline{71.50} \\
    Latent-space Cond.         & \textbf{25.13}    & 5.00           & 71.17 \\
    \bottomrule
  \end{tabular}%
  }
  \vspace{-2mm}
\end{table}

\subsection{Applications and Emerging Capabilities}
\label{sec:applications}
By framing video synthesis as completion on a spatio-temporal canvas, our framework naturally unlocks a diverse set of capabilities beyond the standard tasks evaluated above. As illustrated in Fig.~\ref{fig: teaser}, VideoCanvas supports flexible temporal control at arbitrary timestamps (AnyI2V), arbitrary spatio-temporal patch conditioning (AnyP2V), creative video transitions between non-homologous scenes, long-duration video extension via autoregressive generation, and unified video inpainting, outpainting, and camera control. Detailed results and analysis for each application are provided in Appendix~\ref{app:qualitative} and our project page.

\section{Conclusion}

We formalized the task of arbitrary spatio-temporal video completion and provided a systematic analysis showing that existing conditioning paradigms are structurally limited under causal VAEs. Based on this analysis, we proposed VideoCanvas, a framework that adapts In-Context Conditioning with a hybrid strategy---Spatial Zero-Padding and Temporal RoPE Interpolation---enabling pixel-frame-aware control on frozen VAEs without retraining or new parameters. Experiments on VideoCanvasBench demonstrate state-of-the-art performance across diverse tasks, from multi-frame interpolation to creative video transitions. While our approach scales inference cost with the number of conditioning frames, future work on token pruning and efficient attention could further improve scalability for dense conditioning scenarios.

\section*{Acknowledgements}

This work is partially supported by the National Natural Science Foundation of China (Grant No. 62306261), and The Shun Hing Institute of Advanced Engineering (SHIAE) Grant (No. 8115074). This study was supported in part by the Centre for Perceptual and Interactive Intelligence, a CUHK-led InnoCentre under the InnoHK initiative of the Innovation and Technology Commission of the Hong Kong Special Administrative Region Government. This work is also partially supported by Hong Kong RGC Strategic Topics Grant STG1/E-403/24-N, and CUHK-CUHK(SZ)-GDST Joint Collaboration Fund YSP26-4760949.

{
    \small
    \bibliographystyle{ieeenat_fullname}
    \bibliography{main}
}

\clearpage
\appendix
\section*{Supplementary Material}
\phantomsection
\addcontentsline{toc}{section}{Supplementary Material}
\label{app:supp}

\noindent\textbf{Overview.}
This supplementary material provides additional technical details, extended analyses, and qualitative results for \emph{VideoCanvas}. We also include broader comparisons and ablations that complement the main paper. More results are available on our \textbf{Project Page}: \url{https://onevfall.github.io/project_page/videocanvas/}. The content is organized as follows:

{\small
\begin{itemize}[leftmargin=*, itemsep=2pt, topsep=4pt]
    \item \textbf{Section~\ref{app:relatedwork}} (\hyperref[app:relatedwork]{Related Work}) provides a comprehensive review of related literature.
    \item \textbf{Section~\ref{app:videobackbone}} (\hyperref[app:videobackbone]{Introduction of the Base Text-to-Video Generation Model}) reviews the base text-to-video diffusion backbone and key design choices.
    \item \textbf{Section~\ref{app:impl_details}} (\hyperref[app:impl_details]{Implementation Details}) describes training strategy, compared methods, and evaluation protocols for reproducibility.
    \item \textbf{Section~\ref{supp:benchmark_details}} (\hyperref[supp:benchmark_details]{VideoCanvasBench Construction Details}) details dataset curation, task definitions, and annotation/licensing notes.
    \item \textbf{Section~\ref{app:more_analysis_results}} (\hyperref[app:more_analysis_results]{More Analysis and Results}) provides deeper analyses (e.g., padding robustness) and additional quantitative/qualitative results.
    \item \textbf{Section~\ref{app:qualitative}} (\hyperref[app:qualitative]{Applications and Qualitative Results}) showcases diverse applications and side-by-side visual comparisons across tasks.
\end{itemize}
}
\vspace{0.25em}

\renewcommand{\thetable}{R\arabic{table}}
\renewcommand\thefigure{S\arabic{figure}}

\section{Related Work}
\label{app:relatedwork}

\subsection{Arbitrary Spatio-Temporal Video Completion}

Controllable video generation aims to synthesize content that adheres to user inputs beyond a simple text prompt. Existing approaches are often constrained by rigid, task-specific formats, such as conditioning only on a first frame~\citep{guo2023animatediff, kong2024hunyuanvideo, wan2025wan, shi2024motion, gao2025lora}, on a short initial sequence~\citep{bar2025navigation, yang2025resim}, or on structural inpainting and outpainting~\citep{zhou2023propainter, wang2024your, bian2025videopainter, yang2025gencompositor, he2025inpainting}.
Conceptually, these represent special cases of the broader challenge of video completion, yet prior work has treated them as separate sub-tasks, each requiring specialized solutions.
Several recent works~\citep{voleti2022mcvd,chen2023seine, danier2024ldmvfi} explore aspects of multi-frame conditioning or temporally flexible video generation.
SEINE~\citep{chen2023seine} focuses on short-to-long video transitions and temporal prediction, using a temporal mask modeling objective on continuous frame sequences.
Although it enables multi-frame temporal reasoning, SEINE does not support spatially irregular conditioning.
VDT~\citep{lu2023vdt} takes a complementary approach through masked video modeling on dense spatio-temporal token grids. It relies on a spatial-only VAE (i.e., no temporal compression), producing per-frame token maps over which bi-directional prediction is performed.
MCVD~\citep{voleti2022mcvd} employs masked conditional video diffusion for prediction, generation, and interpolation tasks, but operates on pixel space without leveraging modern causal VAEs.
LDMVFI~\citep{danier2024ldmvfi} applies latent diffusion models specifically to video frame interpolation between given keyframes, but is constrained to dense temporal sequences and does not address arbitrary spatial-temporal positioning.
In contrast, we focus on the task of \textit{arbitrary spatio-temporal video completion} under modern latent video diffusion models with causal video VAEs, addressing the pixel-frame ambiguity challenge that these prior works do not encounter.

\subsection{Paradigms for Video Conditioning}

Achieving arbitrary spatio-temporal control requires a robust conditioning mechanism. Existing approaches can be broadly categorized into three paradigms, each with distinct limitations when applied to our task.
\textit{Latent Replacement}~\citep{hacohen2024ltx, kong2024hunyuanvideo} directly overwrites latent slots with conditional content, but suffers from train-inference mismatch when applied beyond first-frame conditioning, often causing motion collapse.
\textit{Channel Concatenation}~\citep{yang2024cogvideox, wan2025} and adapter-based methods~\citep{mou2024t2i, zhang2023adding, jiang2025vace} fuse conditions via concatenation or lightweight encoders, yet require costly VAE and DiT retraining to handle the zero-padding needed for pixel-frame-aware control.
\textit{In-Context Conditioning (ICC)}~\citep{he2025fulldit2, guo2025long, huang2025unityvideo, yan2025scail, fu2026plenoptic,liu2025context}, pioneered by OminiControl~\citep{tan2024ominicontrol} for images and extended to video by FullDiT~\citep{ju2025fulldit} and UNIC~\citep{ye2025unic}, offers a parameter-free alternative by treating conditions as tokens in a unified sequence. While promising, prior ICC methods struggle with the \textit{pixel-frame ambiguity} introduced by causal VAEs, limiting precise temporal alignment.

Building on ICC, we are the first to enable pixel-frame-level arbitrary spatio-temporal video completion under frozen causal VAEs. Our key innovation is \textit{Temporal RoPE Interpolation}, which assigns fractional temporal positions to conditional tokens, achieving sub-latent precision without VAE retraining. Combined with a hybrid conditioning strategy, our approach unlocks ICC's full potential for fine-grained spatio-temporal control.

\begin{figure}[b]
\centering
\includegraphics[width=\linewidth]{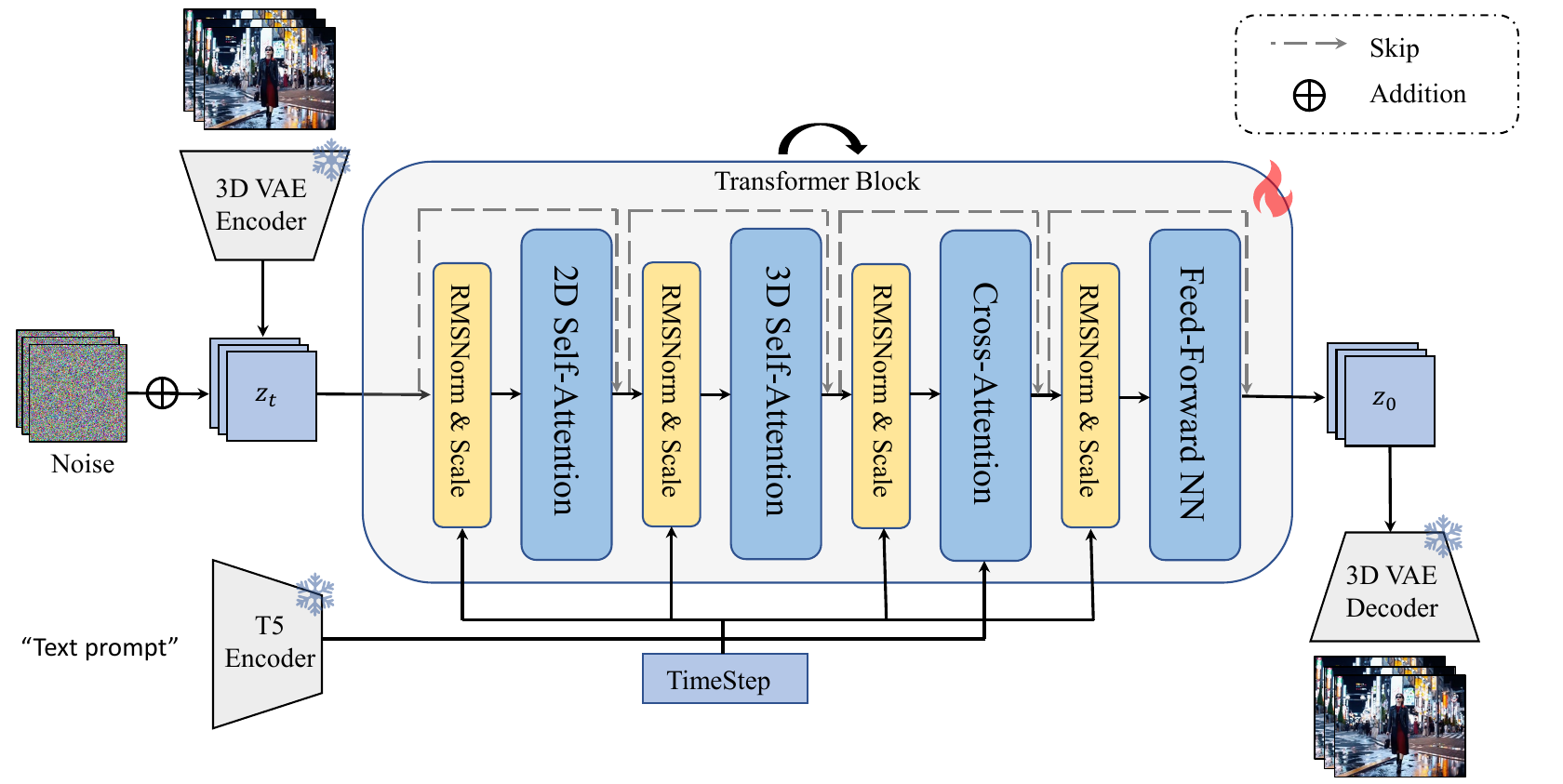}
\caption{\textbf{Overview of the base text-to-video generation model.} }
\label{fig:dit_backbone}
\end{figure}

\section{Introduction of the Base Text-to-Video Generation Model}
\label{app:videobackbone}
We use a transformer-based latent diffusion model ~\citep{dit} as the base T2V generation model, as illustrated in Fig.~\ref{fig:dit_backbone}. We employ a 3D-VAE to transform videos from the pixel space to a latent space, upon which we construct a transformer-based video diffusion model. Unlike previous models that rely on UNets or transformers, which typically incorporate an additional 1D temporal attention module for video generation, such spatially-temporally separated designs do not yield optimal results. We replace the 1D temporal attention with 3D self-attention, enabling the model to effectively perceive and process spatiotemporal tokens, thereby achieving a high-quality and coherent video generation model. Specifically, before each attention or feed-forward network (FFN) module, we map the timestep to a scale, thereby applying RMSNorm to the spatiotemporal tokens.

\section{Implementation Details}
\label{app:impl_details}

\subsection{Training strategy}
The model is fine-tuned for $20$k steps on a curated high-quality video dataset comprising diverse scenes and motion patterns ($384\times672$ resolution, $5$ seconds per clip), using the Adam optimizer with a learning rate of $5\times10^{-5}$ and a batch size of $32$ on $32$ GPUs.

At each iteration, 20 frames are randomly sampled from a source video to serve as temporal anchors.  
From each anchor frame, we extract a spatial region by cropping a patch 
covering between 20\%–100\% of the original frame size.  
This unified training strategy ensures that the model encounters 
a diverse spectrum of conditioning scenarios, ranging from sparse local patches 
to nearly complete frames, and from early anchors to late anchors.  
Such exposure allows the model to learn arbitrary spatio-temporal conditioning in a single framework.

\subsection{Details of Compared Methods}

This section provides additional details on the methods compared against our approach, including (i) existing state-of-the-art models across different tasks, (ii) the conditioning paradigms used for fair evaluation, and (iii) the alignment strategies included in our ablation studies.

\subsubsection{Compared SOTA Methods}

We evaluate our method against state-of-the-art video generation systems spanning multiple task settings:

\begin{itemize}[leftmargin=*]
    \item \textbf{Image-to-Video (I2V).} We compare with CogVideoX-1.5~\citep{yang2024cogvideox} and HunyuanVideo~\citep{kong2024hunyuanvideo}, which represent strong image-to-video generators.
    
    \item \textbf{First–Last-Frame-to-Video (FLF2V).} 
    For FLF2V generation, we include CogVideoX-FT~\citep{CogVideoX-FT} and Sci-Fi~\citep{chen2025sci}, both of which are designed to synthesize temporally coherent sequences conditioned on the first and last frames.
    
    \item \textbf{First–Middle–Last-Frame-to-Video (TF2V).} 
    To evaluate tri-frame conditioning, we decompose the task into two FLF2V subproblems—first$\rightarrow$middle and middle$\rightarrow$last—and stitch the two generated segments into a complete sequence. 
    Both subproblems are solved using CogVideoX-FT~\citep{CogVideoX-FT} and Sci-Fi~\citep{chen2025sci}.

    \item \textbf{First–Last-Patch-to-Video (FLP2V).} Since this task requires completing missing regions from spatial patches, we first convert the patch inputs into full-frame images using FLUX~\citep{flux2024}, and then employ Wan-FLF2V-14B~\citep{wan2025} to generate the full video sequence conditioned on the completed first and last frames.
    
    \item \textbf{Video Inpainting.} For temporal inpainting, we compare with VACE~\citep{jiang2025vace} and ProPainter~\citep{zhou2023propainter} for reconstructing missing or corrupted intervals.
    
    \item \textbf{Video Outpainting.} For spatial and temporal outpainting, we benchmark against VACE~\citep{jiang2025vace} and M3DDM~\citep{fan2023hierarchical}, which are designed to extrapolate beyond the original field of view or temporal extent.
\end{itemize}

\subsubsection{Compared Conditioning paradigms Setting}
For fair comparison of different conditioning paradigms, we re-implement other two representative paradigms on the same base model (Fig.~\ref{fig:intro_figure}b), following the references used in the main text:

\begin{itemize}[leftmargin=*]
    \item \textbf{Latent Replacement}~\citep{hacohen2024ltx,kong2024hunyuanvideo}. 
    For a given conditional frame, the corresponding latent tokens are overwritten 
    with VAE-encoded ground-truth latents. 
    Training applies a masked loss only to non-conditional regions, 
    while conditional regions are assigned timestep 0.
    
    \item \textbf{Channel Concatenation}~\citep{yang2024cogvideox,wan2025}. 
    Condition frames are encoded into latents, assembled into a zero-padded latent sequence, 
    and concatenated with the noisy latent sequence along the channel dimension. 
    A learnable projection layer then restores the embedding dimension. 
    In our implementation, concatenation is applied \textit{after patchification}, 
    as this setting empirically yields the best results; applying it before patchification leads to degraded visual quality. 
    The tradeoff is that after-patchify concatenation substantially increases the channel dimensionality, 
    resulting in a projection layer with $\sim$16.6M trainable parameters. 
    Thus, while this design enriches the conditioning signal and improves learning, 
    it comes at the cost of significantly more parameters compared to the other paradigms.

    \item \textbf{In-Context Conditioning (Ours)}~\citep{tan2024ominicontrol,ju2025fulldit}. 
    Our method encodes condition frames into clean latent tokens and concatenates them 
    with the noisy sequence along the token dimension. 
    Temporal alignment is achieved with our RoPE Interpolation strategy (Sec.~\ref{sec:rope_alignment}). 
    The loss is applied only to noisy tokens, while conditional tokens are assigned timestep 0. 
    This design requires no additional trainable parameters.
\end{itemize}

All paradigms are trained under identical settings and restricted to the same set of conditionable frames 
defined by the VAE stride, ensuring a rigorous and controlled comparison.

\subsubsection{Ablation Alignment Strategies}

\label{app:alignment}

Modern causal video VAEs do not encode frames independently; instead, every latent token represents a temporal window (typically 4 frames).
This design improves compression but introduces a fundamental limitation: a single latent slice does not correspond to a single pixel frame. All compared conditioning strategies differ primarily in how they attempt to recover frame-level alignment from these window-based latents.

\begin{enumerate}[leftmargin=*]
    \item \textbf{Latent-Space Conditioning.} This method encodes the entire groundtruth video with the causal VAE and takes the latent slice whose receptive field overlaps the target frame. However, because each latent mixes information from its surrounding N frames, the “condition latent” inevitably contains content from neighboring frames as well. Thus it does not represent the target frame alone.
    This explains why in Fig.~\ref{fig:psnr_spike} the PSNR peak does not occur at the correct frame index:
    the latent window is misaligned with the temporal position of the condition. Furthermore, conditioning the diffusion model on a temporally mixed latent suppresses motion, producing the collapse observed in Dynamic Degree.

    \item \textbf{Pixel-Space Padding.} This strategy constructs a clip where only the target frame is present and all other frames are zero-padded, then encodes it using the video VAE.
    Zero-padded frames fall inside the VAE’s temporal window, causing the encoder to fuse blank and valid content—an out-of-distribution scenario that leads to color shifts and texture distortion (as shown in Fig.~\ref{fig:padding_vs_rope}).
    Thus, although this method is temporally precise, its reconstructions are of low fidelity.

    \item \textbf{Nearest-slot Assignment (w/o RoPE Interpolation).}
    
    To avoid temporal mixing, we instead encode each conditional frame independently in image mode, which is spatially robust.
    However, image-mode latents are continuous in time, whereas the diffusion transformer expects discrete latent slots determined by the VAE stride.
    Assigning each frame to its nearest temporal slot yields coarse alignment and explains the misaligned PSNR peaks in Fig.~\ref{fig:psnr_spike}.

    \item \textbf{Our Method: Independent Encoding + Temporal RoPE Interpolation.}
    Our approach resolves all issues above.
    By encoding each conditional frame independently, we avoid the temporal entanglement intrinsic to video VAEs.
    Our Temporal RoPE Interpolation then maps each independently encoded latent to an arbitrary continuous temporal coordinate while preserving ordering and positional geometry.
    This provides precise pixel-frame alignment without relying on video-VAE temporal windows or zero-padded inputs.
\end{enumerate}

\subsection{Evaluation Metrics}
For completeness, we provide the full set of evaluation metrics used in our quantitative comparisons. 
In addition to video fidelity measured by FVD~\citep{Unterthiner2018TowardsAG}, we adopt the comprehensive VBench~\citep{huang2024vbench} suite, which evaluates both video quality and temporal consistency across multiple dimensions. 
Specifically, the following metrics are reported:

\begin{itemize}[leftmargin=*]
    \item \textbf{FVD}~\citep{Unterthiner2018TowardsAG} ($\downarrow$): Measures overall video fidelity and temporal coherence with respect to real data.
    
    \item \textbf{Aesthetic Quality}~\citep{LAIONaes} ($\uparrow$): Assesses the perceptual attractiveness of the generated video frames.
    
    \item \textbf{Background Consistency} ($\uparrow$): Evaluates how well the background remains stable across time. This differs from CSCV~\citep{Cai_2025_CVPR} (better for videos with significant camera motion or large scene transition), which measures adjacent-frame CLIP similarity, whereas VBench computes consistency relative to the first frame.

    \item \textbf{Dynamic Degree}~\citep{teed2020raft}($\uparrow$): Quantifies the amount of motion present in the generated video, reflecting whether the dynamics are neither overly static nor excessive.
    
    \item \textbf{Imaging Quality}~\citep{Ke2021MUSIQ} ($\uparrow$): Measures sharpness, clarity, and reconstruction of fine-grained details.
    
    \item \textbf{Motion Smoothness} ($\uparrow$): Captures the temporal stability of motion across adjacent frames.
    
    \item \textbf{Overall Consistency} ($\uparrow$): Evaluates global temporal coherence across the entire clip.
    
    \item \textbf{Subject Consistency} ($\uparrow$): Measures identity and appearance stability of the primary subject across time.
    
    \item \textbf{Temporal Flickering} ($\uparrow$): Detects flickering artifacts or frame-to-frame instability.
    
    \item \textbf{Normalized Average} ($\uparrow$): The mean of all above normalized VBench metrics, providing an aggregated measure of overall video quality.
\end{itemize}

These metrics jointly offer a detailed and reliable characterization of video quality and temporal stability, ensuring a fair and comprehensive comparison across all evaluated methods.

\section{VideoCanvasBench Construction Details}
\label{supp:benchmark_details}

This section provides a comprehensive overview of the data curation and task generation pipeline for \textit{VideoCanvasBench}, the first systematic evaluation suite for arbitrary spatio-temporal video completion.

\subsection{Data Curation}

We curate two complementary types of sources: (1) \textit{homologous} videos for testing fidelity within a single coherent scene, and (2) \textit{non-homologous} images and videos for evaluating creativity across distinct content.

\paragraph{Homologous Video Set (100 Videos).}
We began with an initial pool of $\sim$2,000 videos from Pexels~\citep{pexels2025}. A multi-stage filtering pipeline was applied to ensure quality and diversity:
\begin{itemize}[leftmargin=*]
    \item Blur filtering: 
    blurry videos were removed by calculating the CV2.Laplacian~\citep{opencv_library} score for each frame and excluding those below a threshold of 200.
    \item Motion filtering: static or nearly-static clips were excluded using RAFT-based motion magnitude thresholds exceeding 5~\citep{teed2020raft}.
    \item Length filtering: only videos longer than 5 seconds were retained.
\end{itemize}
From this pool, we selected 100 diverse, high-quality clips covering a wide range of scenes (e.g., human activities, animals, landscapes). All were standardized to 77 frames at 15 FPS to provide a consistent evaluation length. Each video is paired with captions generated by a captioning model fine-tuned on Koala36M~\citep{wang2025koala} following the LLaVA-based~\citep{liu2023visual} annotation pipeline. All captions are further verified by human annotators to ensure accuracy in both content and motion descriptions.

\paragraph{Non-Homologous Image and Video Sets.}
To test the ability to synthesize across unrelated contexts, we manually curated visually distinct sources from Pexels~\citep{pexels2025} and Unsplash~\citep{unsplash2025}, ensuring large appearance and semantic gaps. The set includes:
\begin{itemize}[leftmargin=*]
    \item 50 pairs of non-homologous images, selected to maximize dissimilarity (e.g., indoor vs.\ outdoor, object vs.\ scene).
    \item 50 triplets of non-homologous images, further increasing combinatorial diversity.
    \item 30 pairs of non-homologous video clips, curated for challenging video transitions, similar to the blending function of Sora~\citep{sora}.
\end{itemize}
These non-homologous cases explicitly test the model’s capacity for creative interpolation and cross-scene reasoning. Each non-homologous source is annotated with captions automatically generated by Gemini~2.5 Pro~\citep{comanici2025gemini} and manually corrected to ensure faithful descriptions of both appearance and motion.

\subsection{Benchmark Task Definitions}

\paragraph{Task 1: AnyI2V (Any-Timestamp Image-to-Video).}
This task uses full frames as conditions to test temporal reasoning and interpolation fidelity. 
We explicitly construct nine sub-tasks by combining conditions from fixed temporal anchors: start (frame 1), middle (frame 41), and end (frame 77).
\begin{itemize}[leftmargin=*]
    \item Homologous cases. From each source video we sample three anchor frames (start, middle, end), and construct:
    \begin{itemize}
        \item \textit{Single-frame I2V:} start $\rightarrow$ video, middle $\rightarrow$ video, end $\rightarrow$ video.
        \item \textit{Two-frame I2V:} start+end $\rightarrow$ video, start+middle $\rightarrow$ video, middle+end $\rightarrow$ video.
        \item \textit{Three-frame I2V:} start+middle+end $\rightarrow$ video.
    \end{itemize}
    \item Non-homologous cases. For curated pairs of images, we construct the three two-frame tasks (start+end, start+middle, middle+end). For curated triplets of images, we construct the three-frame task (start+middle+end). Each non-homologous source is annotated with captions automatically generated by Gemini~2.5 Pro~\citep{comanici2025gemini} and manually checked for accuracy.
\end{itemize}

\paragraph{Task 2: AnyP2V (Any-Timestamp Patch-to-Video).}
This variant follows the same nine sub-task definitions as AnyI2V setting, but replaces each full-frame condition with a cropped patch.
\begin{itemize}[leftmargin=*]
    \item Patch extraction. For each conditional frame, patches are obtained via a semi-automated process: 50\% object-aware masks using SAM~\citep{kirillov2023segment} or YOLO~\citep{ultralytics2023yolo}, and 50\% random crops. 
    \item Temporal anchors. The same start, middle, and end frame positions are used to construct single-, two-, and three-frame variants, for both homologous and non-homologous cases. 
    \item Difficulty. The subset explicitly includes challenging cases with very small subjects, requiring the model to extrapolate from minimal context.
\end{itemize}

\paragraph{Task 3: AnyV2V (Transition, Inpainting and Outpainting).}
This task evaluates more general video-level completion scenarios beyond frame- or patch-level control. 
It consists of three sub-categories:

\begin{itemize}[leftmargin=*]
    \item \textit{Video Transition.} For 30 curated pairs of non-homologous video clips, the first clip provides the start segment and the second the end segment, while the model synthesizes the intermediate transition. This setup parallels the blending function explored in Sora~\citep{sora}. Each case is annotated with captions generated by Gemini~2.5 Pro~\citep{comanici2025gemini} and manually corrected to ensure faithful descriptions of both content and motion.
    \item \textit{Inpainting.} For homologous videos, interior rectangular masks are applied to each frame, covering 20\%–50\% of the width/height. The model must fill the missing regions with temporally consistent content. 
    \item \textit{Outpainting.} Boundary masks are applied to crop the central region, masking out 60\%–90\% of the width/height. The model is required to extrapolate plausible outer regions beyond the visible content. 
\end{itemize}

\subsection{Scale}
In total, \textit{VideoCanvasBench} includes over \textbf{2,000} test cases: 900 for AnyP2V, 900 for AnyI2V, and 230 for AnyV2V. Each case is designed to probe a specific aspect of fidelity, creativity, or temporal reasoning in the proposed unified task.

\subsection{Licensing and Annotations.}  
All videos in our benchmark are sourced from Pexels~\citep{pexels2025}, and images are sourced from both Pexels and Unsplash~\citep{unsplash2025}.  
Content on Pexels is provided under the Pexels License, which permits free use for commercial and non-commercial purposes without requiring attribution, with restrictions against reselling unaltered copies, use in trademarks, or misuse of identifiable people or brands.  
A subset of Pexels content is explicitly marked as Creative Commons Zero (CC0), which places the work in the public domain.  
Unsplash photos are provided under the Unsplash License, which similarly allows free commercial and non-commercial use without attribution, while prohibiting resale of unaltered content, creation of competing stock services, or misleading association with brands or people.  
In both cases, all curated data is legally licensed for academic research use.

Captions generated by Gemini~2.5 Pro~\citep{comanici2025gemini} were manually verified by the authors 
to ensure accuracy and consistency across all benchmark cases.

\section{More Analysis and Results}
\label{app:more_analysis_results}
\subsection{Analysis of Zero-Padded Inputs}
\label{app:zero_padding}

\begin{figure*}[ht]
    \centering
    \includegraphics[width=0.9\linewidth]{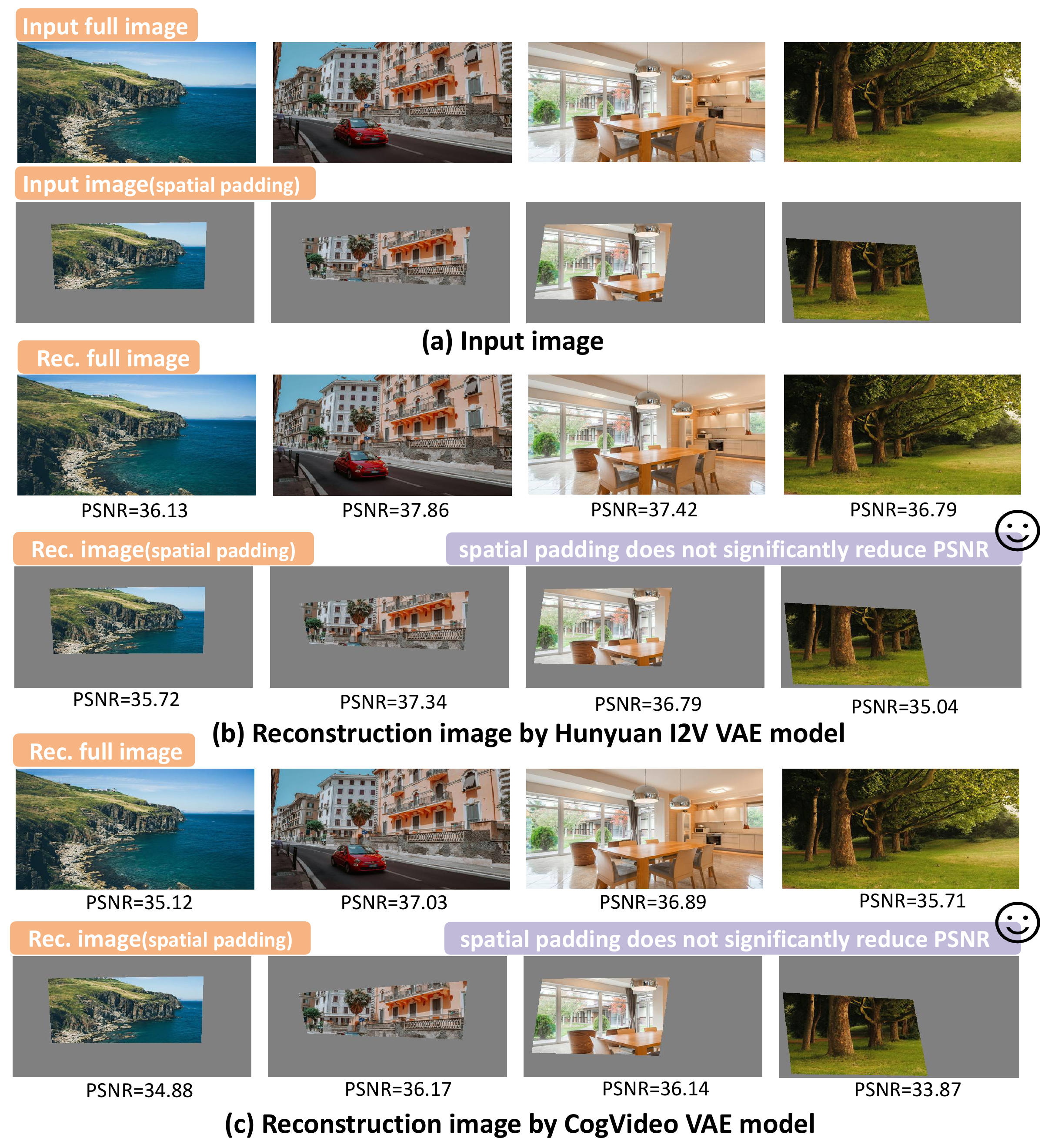}
    \caption{\textbf{Robustness of Hybrid Video VAEs to Spatial Padding}. This figure demonstrates that both the Hunyuan I2V and CogVideo VAE models can tolerate spatial zero-padding well. When reconstructing images with large zero-padded regions (middle row), the PSNR values are only slightly lower than those of the full, unpadded images (top row). Crucially, the original content within the non-zero regions is faithfully preserved, while the padded areas remain visually neutral. This empirical evidence confirms that our spatial conditioning strategy, which relies on zero-padding before VAE encoding, is stable and practical, enabling precise spatial control without degrading the quality of the conditioned content.}
    \label{fig:zero_padding_vis}
\end{figure*}

\begin{figure*}[ht]
    \centering
    \includegraphics[width=0.9\linewidth]{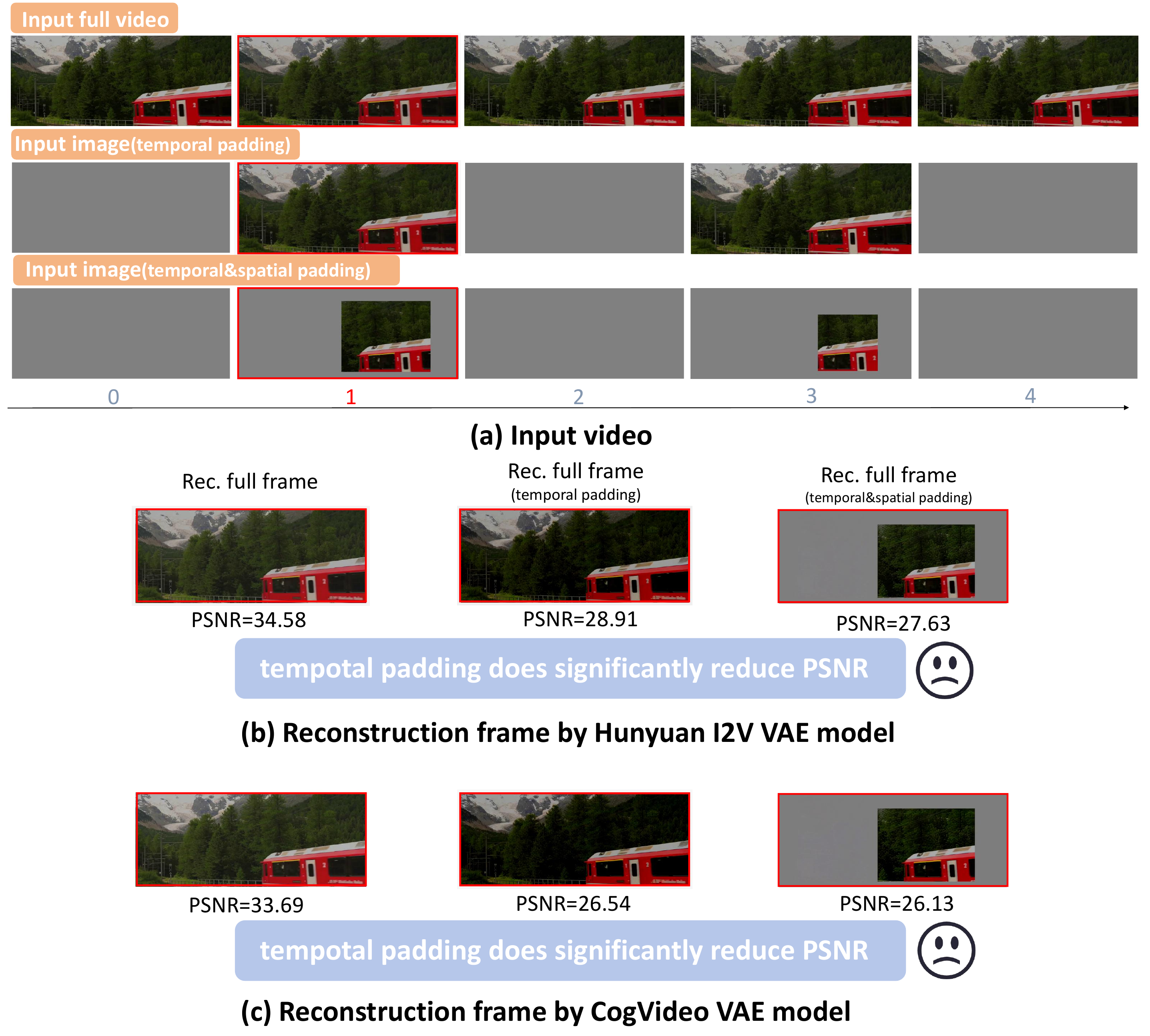}
    \caption{\textbf{Vulnerability of Hybrid Video VAEs to Temporal Padding}. This figure contrasts the robustness observed in spatial padding. When applying temporal zero-padding (where only specific frames contain content), both VAE models suffer a relatively great drop in reconstruction quality. The PSNR values for the padded reconstructions (bottom rows) are much lower than those of the full video (top row), demonstrating a degradation in fidelity. The reconstructed frames exhibit noticeable color shifts, and loss of detail, highlighting that the VAE cannot handle such distributionally mismatched inputs. This mode underscores why direct temporal zero-padding is ineffective and validates the necessity of our Temporal RoPE Interpolation strategy, which avoids this problem by operating at the latent token level with fractional positions.}
    \label{fig:temporal_padding_vis}
\end{figure*}

In Section~\ref{sec:method}, we describe using zero-padding to indicate unconditioned regions when preparing conditional frames. This approach is crucial for our spatial conditioning strategy, as it allows us to precisely specify the location of a condition patch within a frame without modifying the pre-trained VAE backbone. However, a critical question arises: can a standard hybrid video VAE, trained on natural images and videos, effectively handle inputs that contain large areas of zero-valued pixels (i.e., spatial padding)? As illustrated in Figure~\ref{fig:zero_padding_vis} and Figure~\ref{fig:temporal_padding_vis}, this distinction between spatial and temporal padding is fundamental to understanding our method.

To address this, we conducted an empirical study using two popular pre-trained VAE models: Hunyuan I2V and CogVideo. We evaluated their robustness to both spatial and temporal padding under realistic conditions.

\paragraph{Setup.}
We collected 20 diverse full-resolution images and 20 short video clips from YouTube, representing a wide range of content (e.g., landscapes, cityscapes, indoor scenes, moving vehicles). For each image, we applied random spatial zero-padding masks, covering approximately 40-60\% of the pixels. For each video clip, we created three types of padded inputs:
1.  A video with conditional frames containing the original content, while all other frames are filled with zeros (pure temporal padding).
2.  A video where conditional frames contains cropped region of the original content, with all other frames being zero (temporal \& spatial padding).

Each input was then encoded and decoded using the two hybird VAE model. We measured the reconstruction fidelity using PSNR and qualitatively inspected the outputs.

\paragraph{Reconstruction Results.}
The results provide clear evidence of the differential impact of padding modes:

Spatial Padding Robustness: As shown in Figure~\ref{fig:zero_padding_vis}, both VAE models demonstrate remarkable tolerance to spatial zero-padding. The average PSNR of reconstructed images with spatial padding is only marginally lower than that of the baseline (full image), with an average drop of \textbf{0.89 dB}(Hunyuan I2V) and \textbf{1.13 dB}(CogVideo).

Temporal Padding Vulnerability: In stark contrast, Figure~\ref{fig:temporal_padding_vis} reveals the limitations of traditional approaches. When applying temporal zero-padding (encoding a single frame into a sequence where most frames are zero), both VAE models exhibit a dramatic degradation in reconstruction quality. The average PSNR drops by over \textbf{6.12 dB}(Hunyuan I2V) and \textbf{7.01 dB}(CogVideo) compared to the baseline. 

\begin{figure*}[t]
    \centering
    \includegraphics[width=0.9\linewidth]{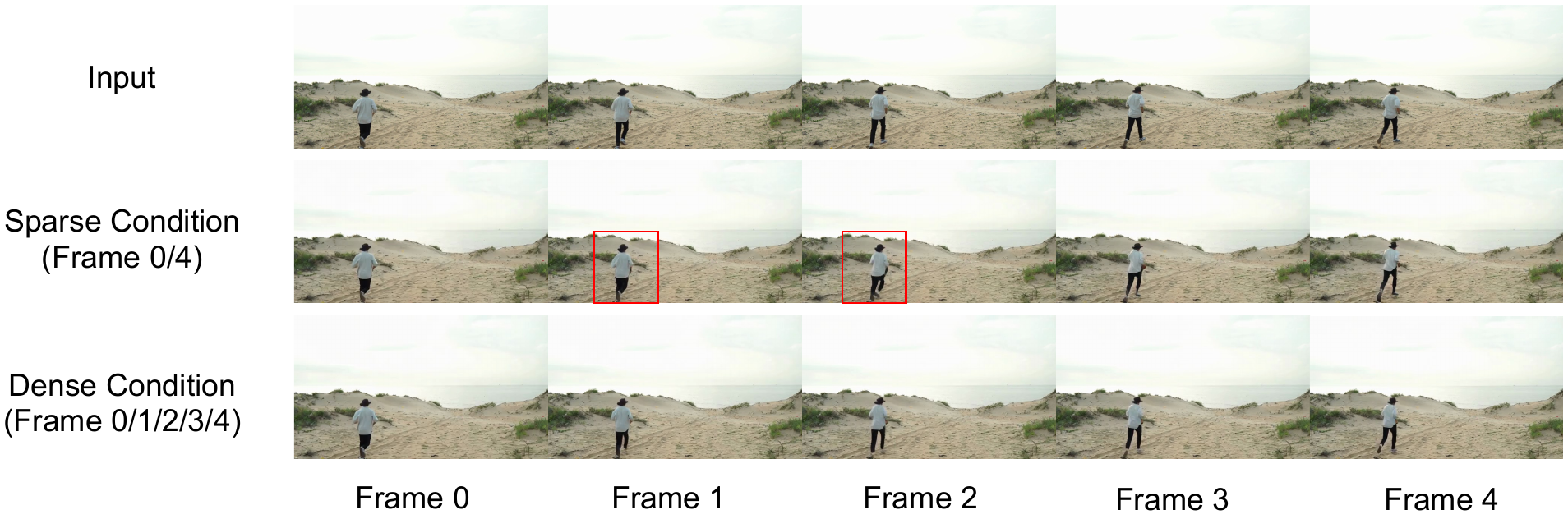} %
    \caption{
      \textbf{Visual comparison of Sparse vs. Dense conditioning.}
      The top row shows the ground-truth frames. The middle row (\textbf{Sparse}) is generated using only frames 0 and 4 as conditions; note the plausible but slightly drifted interpolation for the intermediate frames. The bottom row (\textbf{Dense}) is generated using all five frames as conditions, resulting in a near-perfect reconstruction. This highlights the benefit of dense, frame-by-frame control—a capability unique to our method.
      }
    \label{fig:vis_dense_sparse}
\end{figure*}

\paragraph{Conclusion.}
These findings confirm that the key to achieving pixel–frame-aware control lies in decoupling spatial and temporal handling. Our method leverages the inherent robustness of the VAE to spatial padding while bypassing the ineffectiveness of temporal padding through our proposed Temporal RoPE Interpolation. This separation enables flexible, high-fidelity video completion using a frozen VAE without requiring retraining or architectural modification. 

In addition to the controlled analyses presented above, our qualitative results under arbitrary spatiotemporal conditioning (Fig.~\ref{fig: teaser} and Fig.~\ref{fig:qual_anyp2v}) further demonstrate that spatial zero-padding remains stable even under large variations in placement, content, and temporal context. These observations provide complementary evidence supporting the effectiveness and general applicability of our approach across diverse zero-padded settings. Together with the quantitative results reported in the main text, these findings consistently validate the necessity and effectiveness of our design.

\subsection{Advantages of Temporal RoPE Interpolation}
\label{app:rope}

Figure~\ref{fig:psnr_spike} in the main paper has shown that our Temporal RoPE Interpolation achieves \textit{precise one-to-one alignment} between condition frames and their target temporal positions. Here we further demonstrate not only that our model can leverage this precision for \textit{dense} conditioning, but also why this capability represents a crucial advantage over competing paradigms.

To this end, we conduct an additional experiment on the homologous video set from  \textit{VideoCanvasBench}. Each 77-frame video is conditioned on the first five frames (0-4) in two different ways:

\begin{itemize}[leftmargin=*]
    \item \textbf{Sparse Condition:} Only the boundary frames (0 and 4) are provided. The model must interpolate the three missing frames (1, 2, 3) in between.
    \item \textbf{Dense Condition:} All five frames (0, 1, 2, 3, 4) are explicitly provided as conditions, testing frame-wise alignment at every step.
\end{itemize}

Both settings are used to generate the full video, and we evaluate the fidelity by computing PSNR on the first 5 frames against the ground truth.

\begin{table}[h!] %
\centering
\caption{Average PSNR (dB) across 100 videos under sparse vs. dense conditioning.}
\label{tab:sparse_dense_psnr}
\begin{tabular}{lcc}
\toprule
Condition Type & Conditioned Frames & PSNR (↑) \\
\midrule
Sparse (two frames) & 0, 4 & 24.789 \\
Dense (five frames) & 0, 1, 2, 3, 4 & \textbf{25.033} \\
\bottomrule
\end{tabular}
\end{table}

The quantitative results in Table~\ref{tab:sparse_dense_psnr} confirm that explicitly conditioning on consecutive frames yields higher reconstruction fidelity. Figure~\ref{fig:vis_dense_sparse} provides a visual illustration. In the sparse case, our model generates a plausible interpolation, but with minor, expected drift in the unconditioned intermediate frames. The dense case, in contrast, achieves a near-perfect reconstruction.

This comparison highlights a fundamental limitation of paradigms like \textit{Channel Concatenation}. Due to their coarse, slot-based nature and the constraint of a frozen VAE, they can only condition on one frame per latent slot (e.g., one frame for every $N=4$ pixel frames). They are therefore structurally incapable of providing dense guidance for the intermediate frames (e.g., frames 1, 2, 3) and are locked into a "sparse" conditioning mode, inevitably suffering from the kind of interpolation drift shown in our sparse example. In contrast, our \textit{Temporal RoPE Interpolation} uniquely enables true dense conditioning, allowing VideoCanvas to maintain high fidelity frame-by-frame—a capability that is structurally inaccessible to these competing methods.

\subsection{More Ablation and User Studies}
\label{supp:ablation}

\subsubsection{Ablation of RoPE Strategy}
\label{supp:ablation_rope}

\begin{figure}[ht]
\centering
\includegraphics[width=1\linewidth]{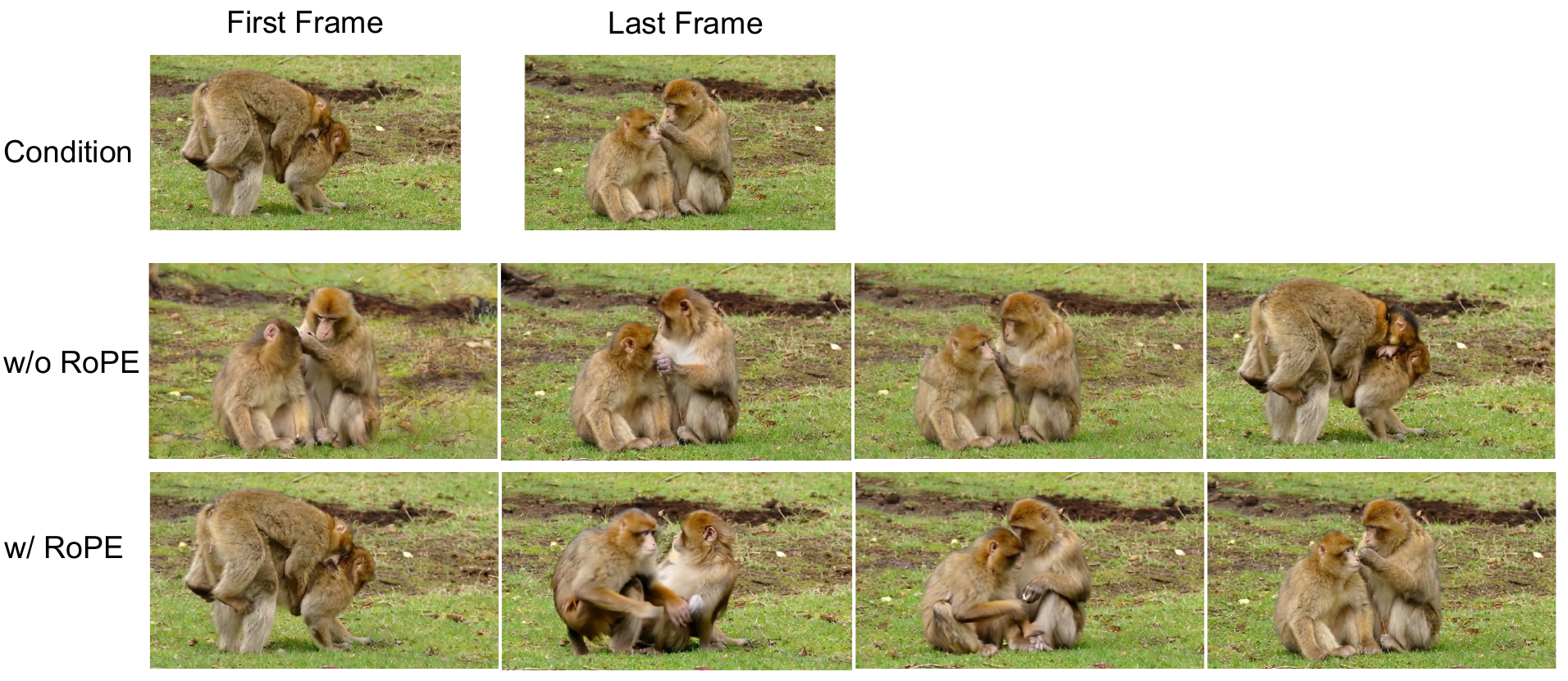}
\vspace{-6pt}
\caption{Visualization of without RoPE.}
\label{fig:no_rope}
\end{figure}

We further investigate the impact of different Rotary Positional Embedding (RoPE) variant strategies on temporal alignment. Specifically, we compare our proposed \textbf{Fractional Interpolation} (mapping frames to $0, 0.25, \dots, 19$) against an \textbf{Integer Extrapolation} strategy (scaling indices to $0, 1, \dots, 76$ for a $4\times$ expansion) and a baseline \textbf{w/o RoPE}.

\begin{table}[h]
    \centering
    \small
    \caption{\textbf{Quantitative ablation of RoPE strategies.} We report the FVD ($\downarrow$) scores (lower is better) on the validation set. Our fractional interpolation strategy outperforms both integer extrapolation and the baseline without RoPE.}
    \label{tab:rope_ablation}
    \begin{tabular}{l c c}
        \toprule
        Method & Index Mapping Strategy & FVD ($\downarrow$) \\
        \midrule
        w/o RoPE & N/A & 12.568 \\
        Integer Extrapolation & $0, 1, \dots, 76$ & 11.079 \\
        \textbf{Ours (Fractional)} & $\mathbf{0, 0.25, \dots, 19}$ & \textbf{10.943} \\
        \bottomrule
    \end{tabular}
\end{table}

As reported in Table~\ref{tab:rope_ablation}, our method achieves the best performance with the lowest FVD score. We attribute the superiority of our fractional strategy to the preservation of pre-trained priors. Since the base model was trained on a specific temporal range (indices $0 \sim 19$), our fractional approach ensures the input indices remain within this learned distribution. This allows the model to effectively inherit the original temporal priors, leading to faster convergence and better performance compared to the Integer method, which shifts the distribution to an unfamiliar range ($0 \sim 76$).

Furthermore, removing RoPE proves disastrous. Without explicit positional indicators, the condition tokens injected via concatenation lack the temporal cues necessary to ``lock'' onto specific frame positions. Consequently, the mechanism degenerates from precise conditioning into a loose ``image reference'' mode, where the generated frames fail to strictly adhere to the condition frames. This degradation is visually evident in Fig.~\ref{fig:no_rope}, where the model fails to maintain temporal position consistency.

\subsubsection{User Study on Conditioning Paradigms}

To complement the quantitative results in Table~\ref{tab:paradigm_comparison}, where our method achieves the best FVD and Dynamic Degree, we further conducted a human preference study using the diverse tasks defined in the VideoCanvas benchmark. We compare our \textbf{In-Context Conditioning (ICC)} against the two representative paradigms: \textit{Latent Replacement} and \textit{Channel Concatenation}.

\textbf{Setup.}
We sampled generated videos across the broad range of tasks in VideoCanvas. Evaluators performed pairwise comparisons based on visual quality, motion smoothness, and condition fidelity.

\textbf{Results.}
As shown in Table~\ref{tab:user_preference}, our method consistently outperforms both baselines:
\begin{itemize}[leftmargin=*]
\item \textbf{Vs. Latent Replacement:} Ours achieves a \textbf{65.4\%} win rate. Users penalized the replacement method for its lower dynamic degree (consistent with Table~\ref{tab:paradigm_comparison}) and disjointed motion transitions.
\item \textbf{Vs. Channel Concatenation:} Ours wins by \textbf{48.2\%} (with a significant tie rate). This confirms that our parameter-free ICC strategy is more effective than increasing channel dimensionality (\~16.6M extra params), yielding better visual harmony.
\end{itemize}
These results align with the quantitative metrics in Table~\ref{tab:paradigm_comparison}, verifying that ICC provides the most effective guidance for the diffusion model.

\begin{table}[h]
\centering
\caption{\textbf{Human Preference Evaluation.} Comparison of our In-Context Conditioning (Ours) against two representative paradigms: \textit{Latent Replacement} and \textit{Channel Concatenation}. We report the percentage of user votes favoring our model versus the variants, averaged across all evaluated tasks on the VideoCanvas benchmark.}
\label{tab:user_preference}
\resizebox{0.95\linewidth}{!}{%
\begin{tabular}{l|ccc}
\toprule
\multirow{2}{*}{Comparison} & \multicolumn{3}{c}{Preference Rate (Ours vs. Variant)} \\
\cmidrule(l){2-4} 
 & Win ($\uparrow$) & Tie & Loss ($\downarrow$) \\ 
\midrule
Ours vs. Latent Replacement & \textbf{65.4\%} & 22.1\% & 12.5\% \\
Ours vs. Channel Concatenation & \textbf{48.2\%} & 28.3\% & 23.5\% \\
\bottomrule
\end{tabular}%
}
\end{table}

\subsection{Training and Inference Cost}
\label{supp:cost}

\begin{table}[!htbp]
\centering
\caption{Training and inference cost comparison across paradigms. 
Training time is measured over 20k steps. 
Inference time is per 77-frame video at $384\times672$ with different numbers of conditional frames.}
\label{tab:cost}

\resizebox{\linewidth}{!}{
\begin{tabular}{lccccc}
\toprule
Method & New Params & Train & \multicolumn{3}{c}{Inference} \\
\cmidrule(lr){4-6}
 &  &  & 1 frame & 2 frame & 3 frame \\
\midrule
Replacement & 0 & 21.47h & 159s & 159s & 159s \\
Channel Concat & \textbf{16.6M} & 22.47h & 164s & 164s & 164s \\
ICC (Ours) & 0 & 24.54h & 168s & 175s & 184s \\
\bottomrule
\end{tabular}
}
\end{table}

Tab.~\ref{tab:cost} compares the computational cost of different conditioning paradigms. Unlike channel concatenation, which relies on a 16.6M projection layer, our ICC design introduces no additional parameters, and the training cost remains comparable (24.5h vs.~21–22h) since ICC only adds lightweight spatio-temporal tokens.

During inference, ICC exhibits a content-aware and controllable scaling: the compute grows with the number of conditioning frames, because a richer conditioning context requires processing longer input sequences within the transformer. For a 77-frame video at $384\times672$, inference takes 168 s with a single conditioning frame, and gradually increases to 333 s / 520 s / 910 s when conditioning on 20 / 40 / 77 frames, respectively.

In sparse conditioning scenarios, the additional cost is almost negligible, since only a few frames expand the sequence length. When more frames are provided, ICC allows users to intentionally trade additional computation for higher fidelity on the corresponding conditioned timestamps, offering stronger control instead of incurring unavoidable overhead. While this scaling makes ICC marginally slower than baselines with fixed-cost inference, the trade-off is justified, as ICC consistently yields higher fidelity and better spatio-temporal alignment (see Sec.~\ref{sec:main_results}, Tab.~\ref{tab:main_benchmark_results} and Tab.~\ref{tab:paradigm_comparison}).

This cost profile is orthogonal to recent acceleration-oriented designs. For example, FullDiT2~\citep{he2025fulldit2} improves ICC efficiency through token selection and context caching, while SURF~\citep{ding2026surf} targets fast high-resolution video generation by preserving the signature of a pretrained generator. These directions could be integrated with our pixel-frame-aware conditioning to further reduce latency without changing the core alignment mechanism.

\section{Applications and Qualitative Results}
\label{app:qualitative}

\subsection{Applications}

The teaser figure (Fig.~\ref{fig: teaser}) has shown some cases, and in this section, we provide extensive qualitative results to demonstrate the versatility and effectiveness of our VideoCanvas framework across a wide range of applications. 

\begin{figure*}[ht]
\centering
\includegraphics[width=0.7\linewidth]{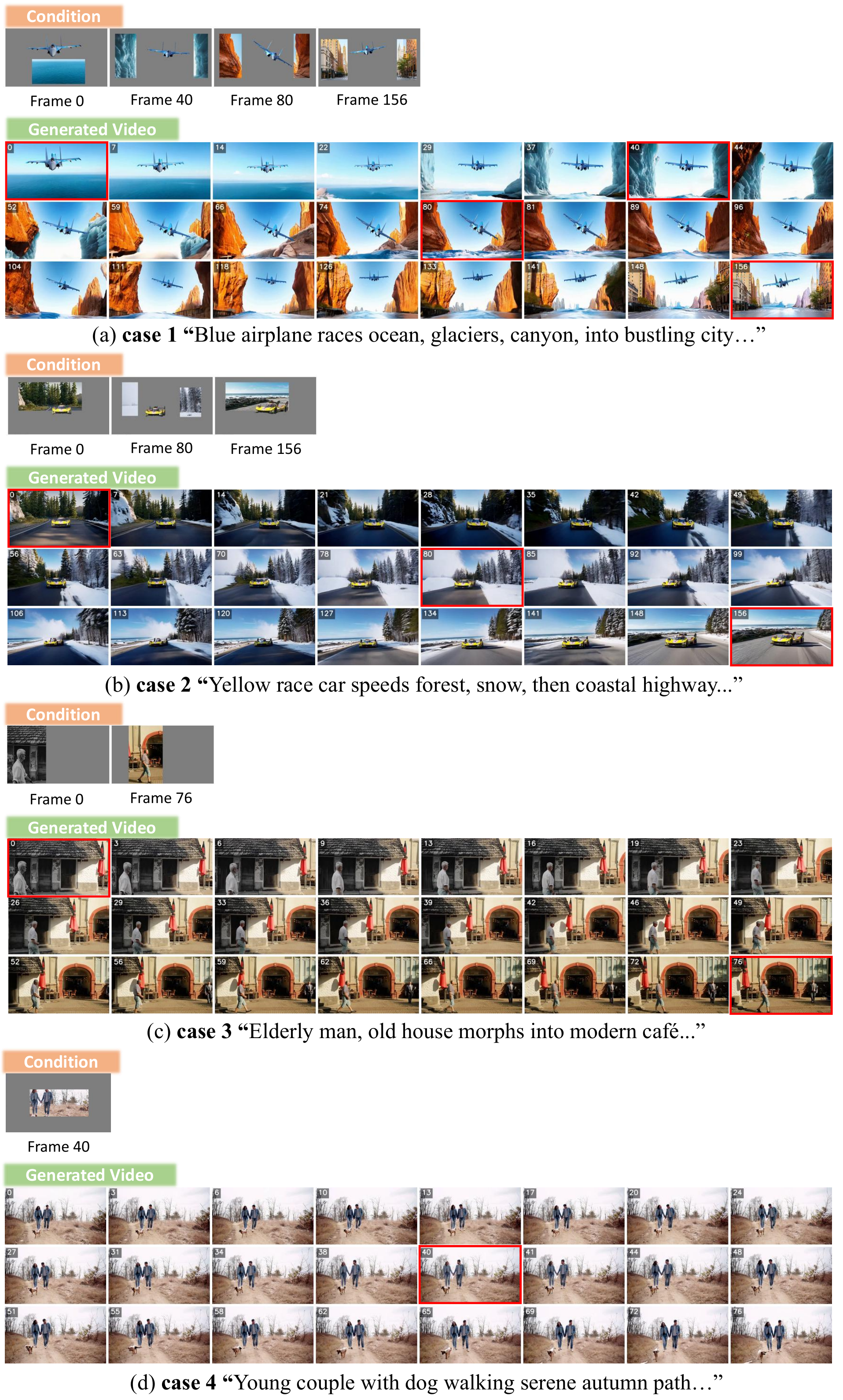}
\caption{\textbf{Results on Any-timestamp Patches to Videos.}}
\label{fig:qual_anyp2v}
\end{figure*}

\begin{figure*}[ht]
\centering
\includegraphics[width=0.7\linewidth]{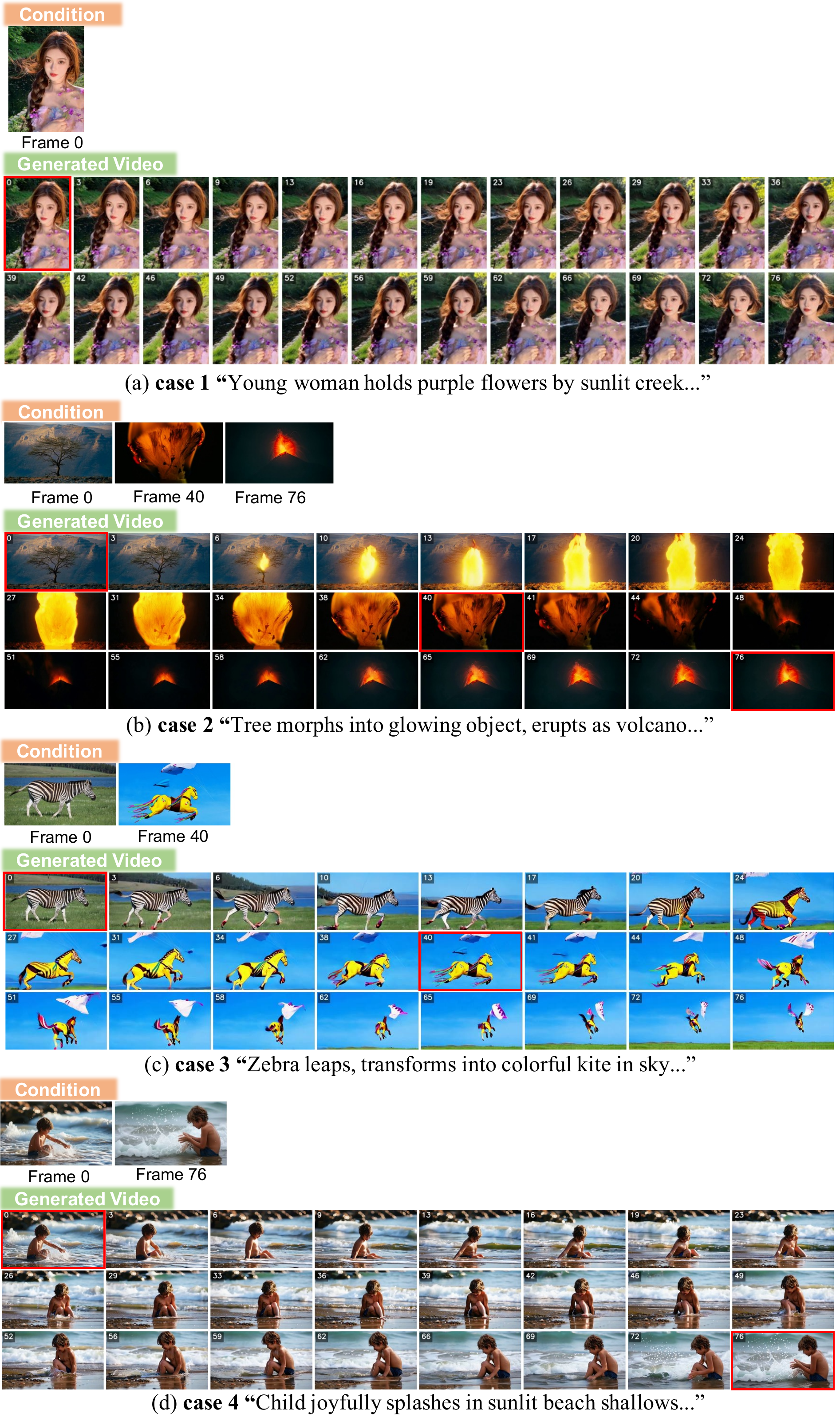}
\caption{\textbf{Results on Any-timestamp Images to Videos.}}
\label{fig:qual_anyi2v}
\end{figure*}

\begin{figure*}[ht]
\centering
\includegraphics[width=0.7\linewidth]{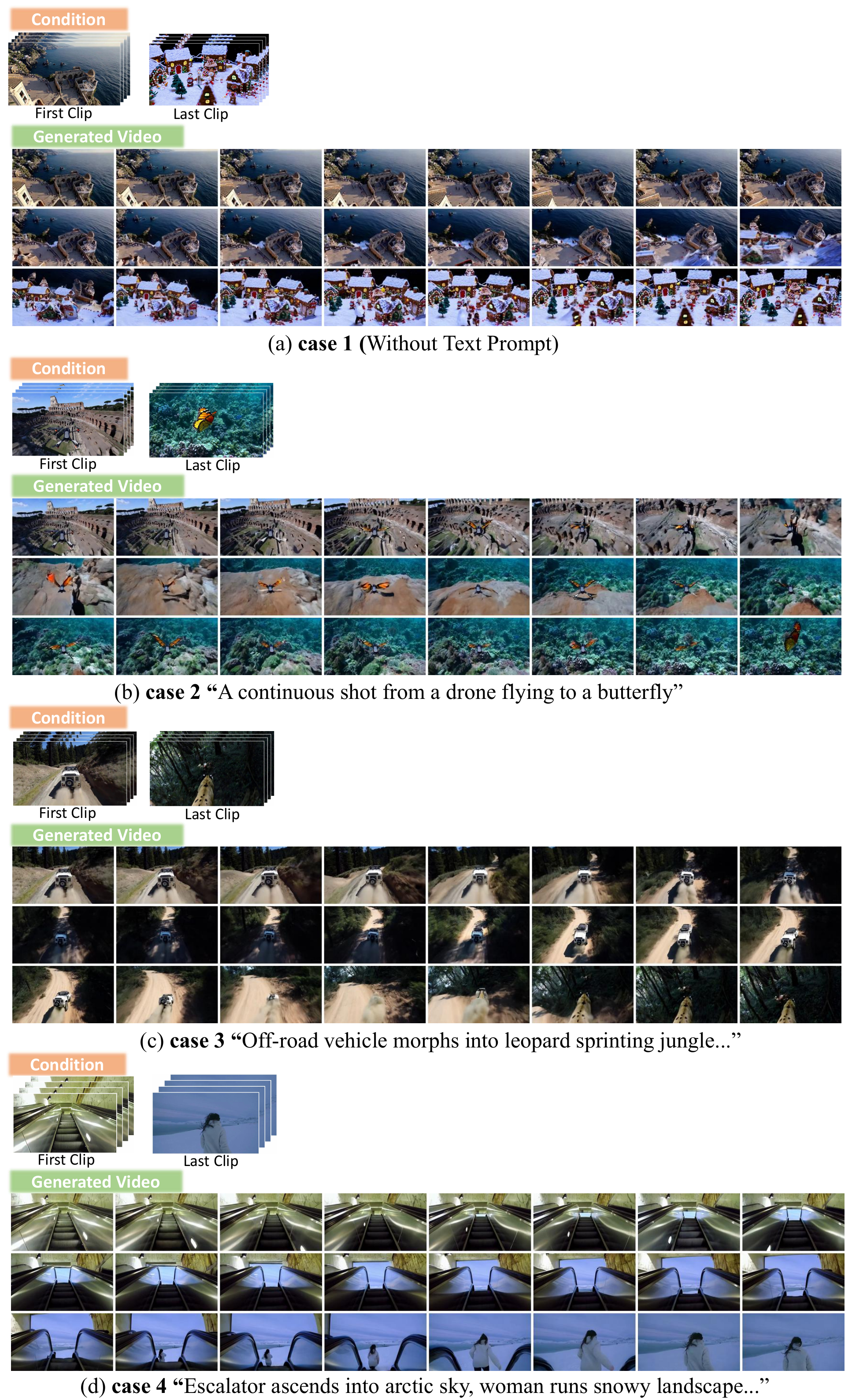}
\caption{\textbf{Results on Video Transition.}}
\label{fig:qual_vt}
\end{figure*}

\begin{figure*}[ht]
\centering
\includegraphics[width=0.85\linewidth]{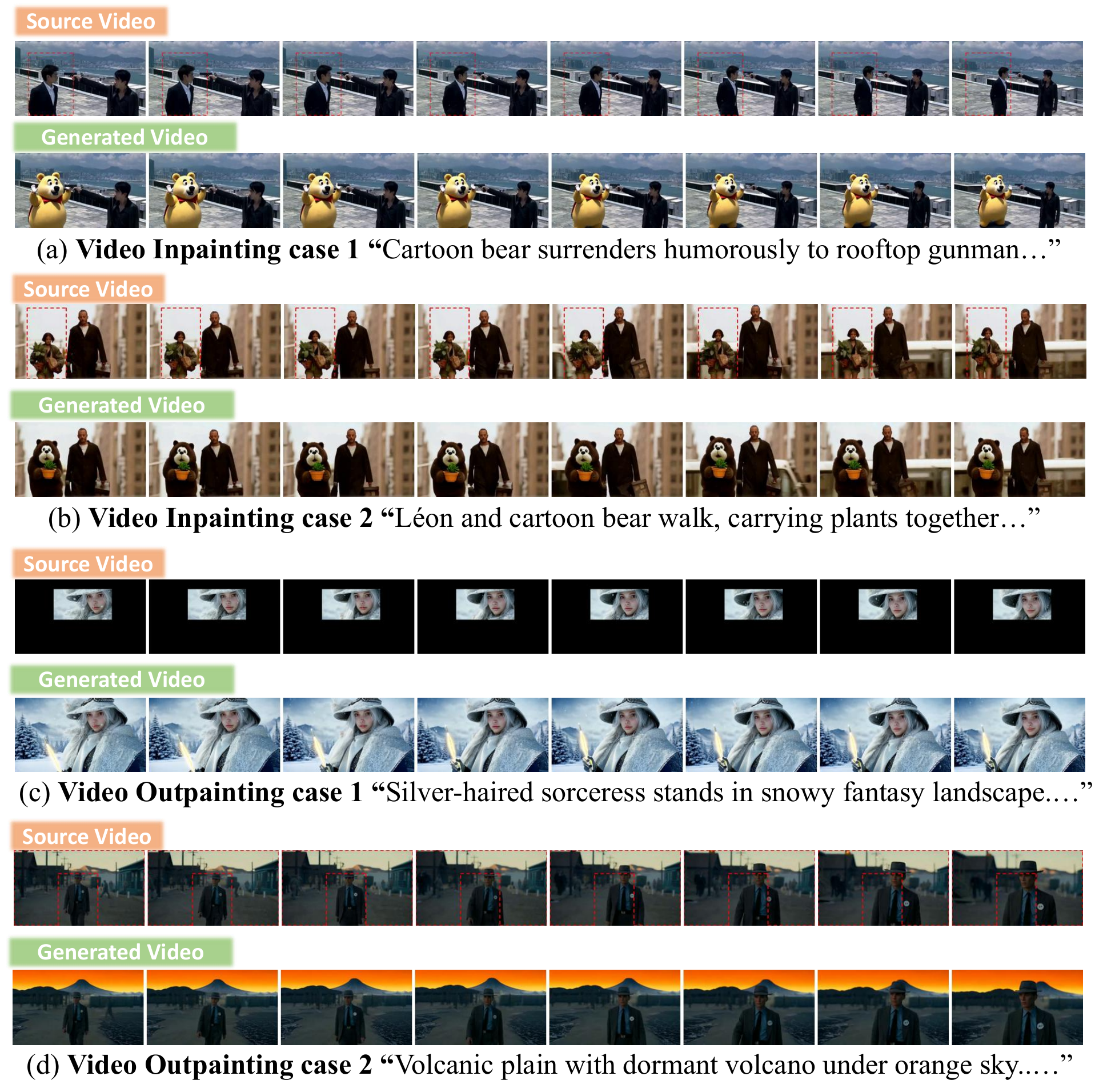}
\caption{\textbf{Results on Video Inpainting and Outpainting.} The red dashed contours indicate the regions that are subject to inpainting or outpainting.}
\label{fig:qual_vivo}
\end{figure*}

\begin{figure*}[ht]
\centering
\includegraphics[width=0.85\linewidth]{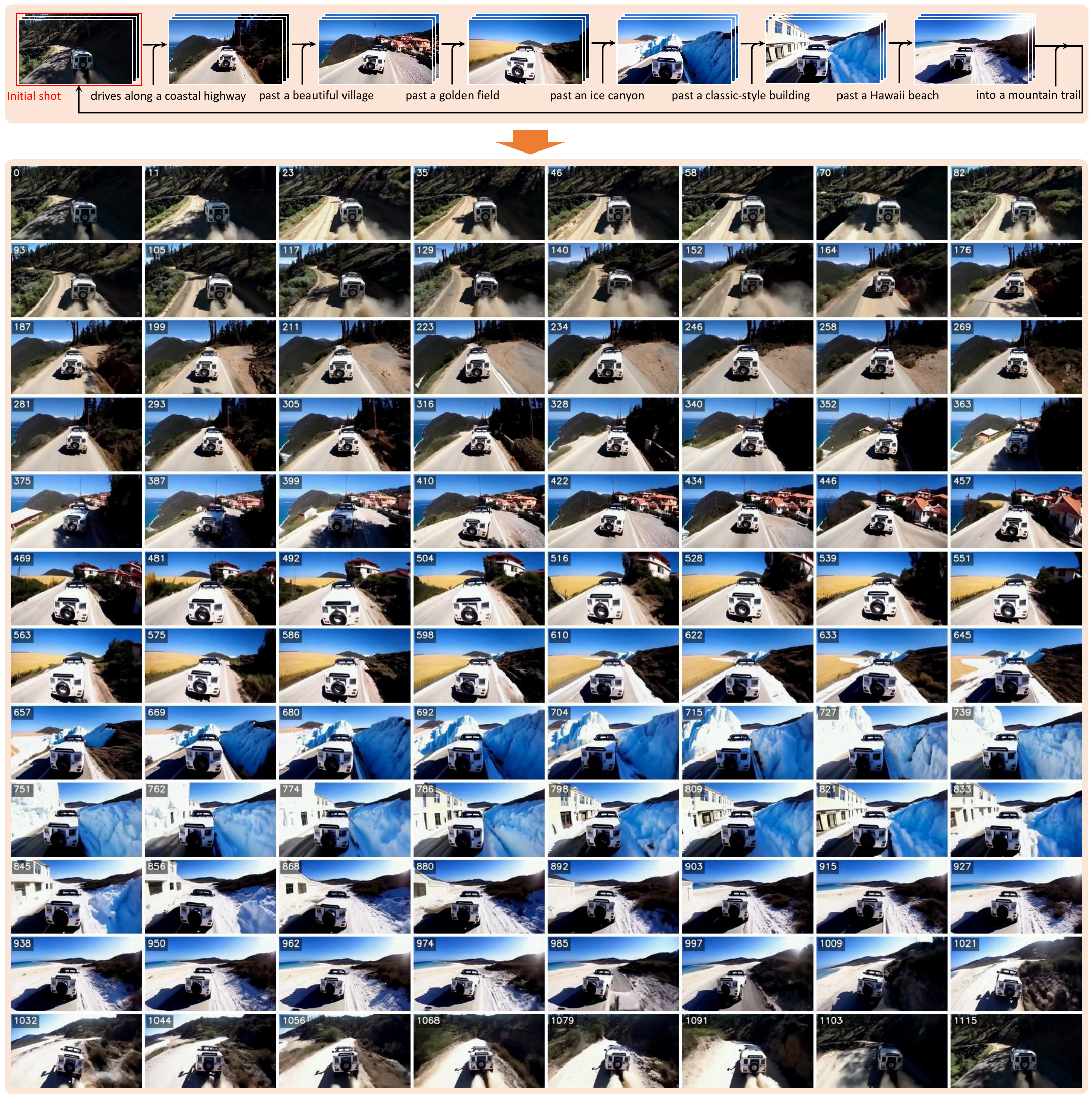}
\caption{
  \textbf{Results on Video Extension and Seamless Looping.} 
  The example showcases a video extended to \textbf{over 1,000 frames} by first applying our video extension capability and then generating a seamless transition back to the initial state. This highlights our model's ability to maintain temporal consistency and visual quality over a long generation horizon without suffering from quality degradation or motion collapse.
}
\label{fig:qual_ve}
\end{figure*}

\begin{figure*}[ht]
\centering
\includegraphics[width=0.8\linewidth]{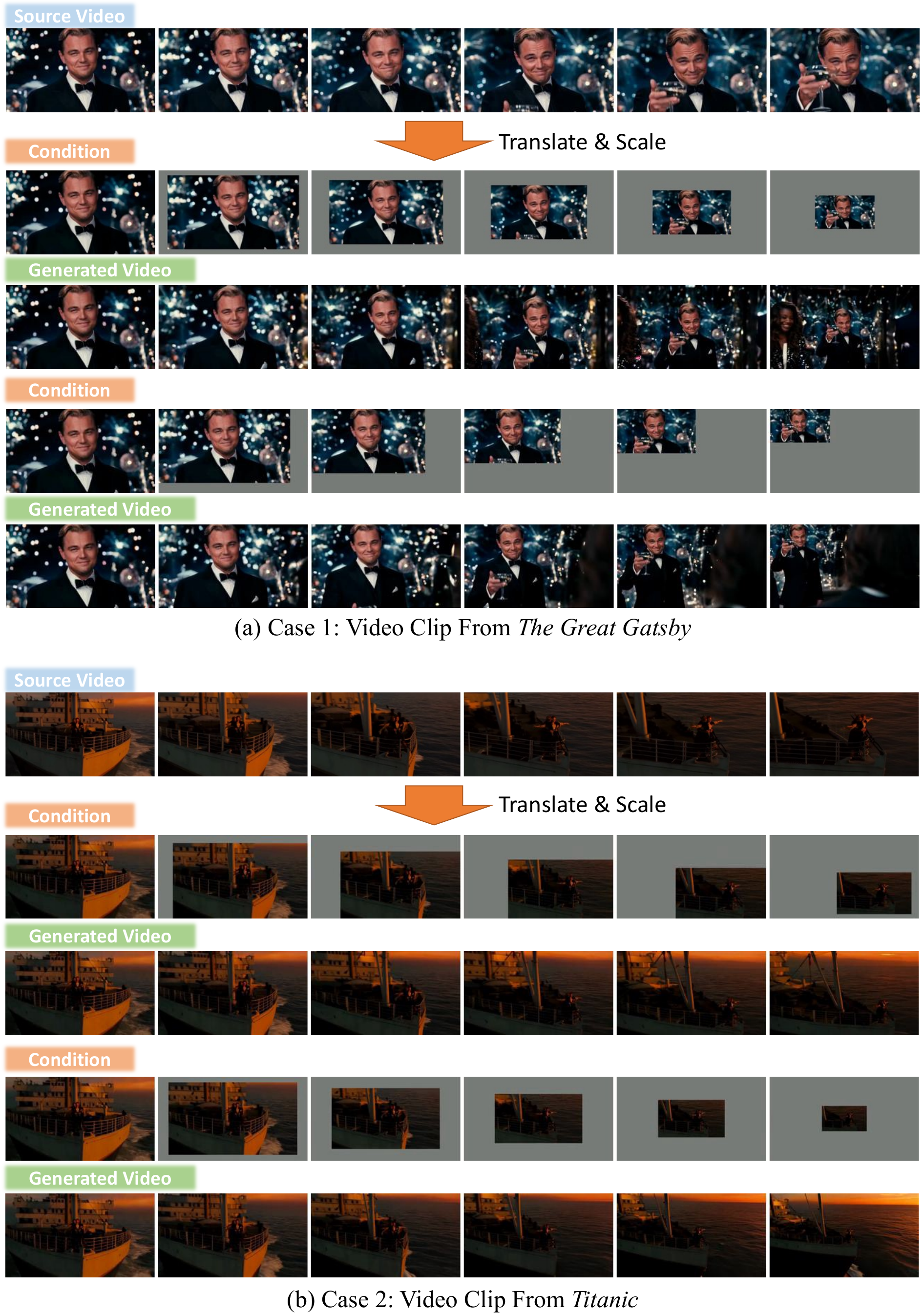}
\caption{
  \textbf{Results on Video Camera Control.} The examples showcase emulated camera effects such as zoom and pan, achieved by progressively translating and scaling content on the spatio-temporal canvas.
}
\label{fig:qual_vc}
\end{figure*}

\begin{itemize}[leftmargin=*]
    \item \textbf{Any-Timestamp Patch-to-Video (AnyP2V).} In Figure~\ref{fig:qual_anyp2v}, we demonstrate our core capability of generating a complete video from a varying number of sparse patches. We showcase challenging scenarios using one, two, three, and even four conditional patches, placed at arbitrary timestamps to rigorously test the model's spatio-temporal reasoning beyond simple first-frame conditioning.

    \item \textbf{Any-Timestamp Image-to-Video (AnyI2V).} Figure~\ref{fig:qual_anyi2v} illustrates the flexibility of our framework on full-frame conditions. The examples include standard cases like first-frame I2V and first-last-frame interpolation, as well as more challenging scenarios where conditions are placed at arbitrary middle timestamps, a capability not well supported by prior methods.

    \item \textbf{Video-Level Completion and Creation (AnyV2V).} Our framework naturally unifies a variety of video editing and creation tasks within a single model. We provide examples of:
    \begin{itemize}
        \item \textbf{Video Transition:} Creative transitions between non-homologous clips are demonstrated in Figure~\ref{fig:qual_vt}.
        \item \textbf{Video Painting:} Inpainting and outpainting results are shown in Figure~\ref{fig:qual_vivo}, where the red dashed contours indicate the generated regions.
        \item \textbf{Video Extension and Looping:} As demonstrated in Figure~\ref{fig:qual_ve}, we showcase long-duration synthesis by extending short clips to over a minute in length while maintaining temporal consistency. This capability can be guided by interactive text prompts to evolve the narrative. Furthermore, we can create perfectly seamless loops by generating a smooth transition from the video's end back to its beginning. Our approach leverages motion context from the last segment's frames to effectively avoid the stuttering artifacts that are common in naive first-last frame-looping methods.

        \item \textbf{Video Camera Control:}  As demonstrated in Figure~\ref{fig:qual_vc}, our framework can emulate camera cinematography by progressively translating or scaling content on the spatio-temporal canvas. This enables a variety of standard camera effects, such as zooms and pans. We showcase this capability by applying dynamic camera movements to classic movie shots, demonstrating its potential for creative post-production.

    \end{itemize}
\end{itemize}

\subsection{Qualitative Comparison}

The following figures showcase side-by-side comparisons with baseline paradigms, illustrating our method's superior performance in motion smoothness, detail sharpness, and temporal consistency.

\begin{figure*}[ht]
\centering
\includegraphics[width=0.9\linewidth]{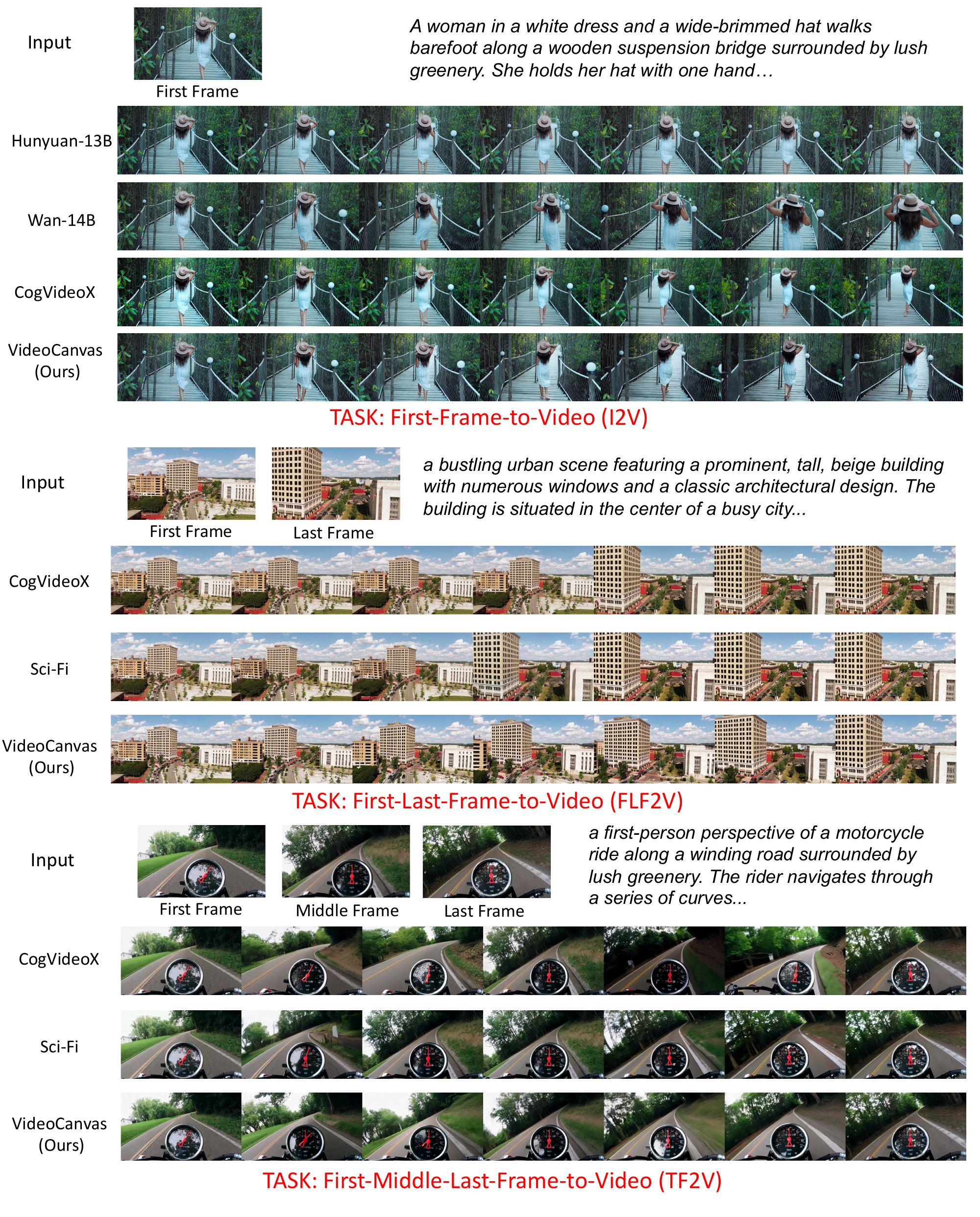}
\vspace{-6pt}
\caption{\textbf{Comparisons with baseline models (1/2).} }
\label{fig:baseline_comparison_1}
\end{figure*}

\begin{figure*}[ht]
\centering
\includegraphics[width=0.9\linewidth]{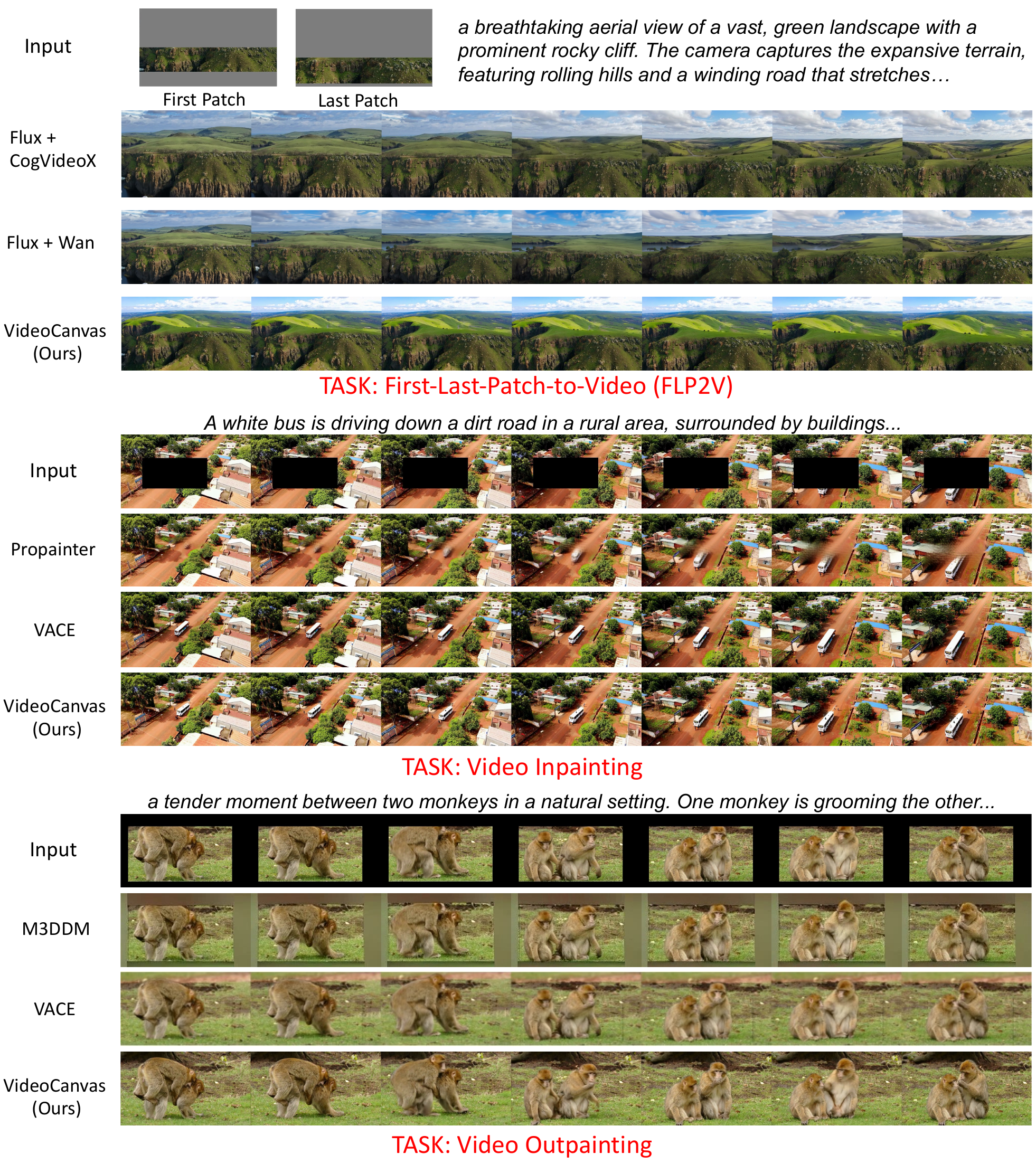}
\caption{\textbf{Comparisons with baseline models (2/2).} }
\label{fig:baseline_comparison_2}
\end{figure*}

Fig.~\ref{fig:baseline_comparison_1} and Fig.~\ref{fig:baseline_comparison_2} present qualitative results across the six varied tasks (I2V, FLF2V, TF2V, FLP2V, Inpainting, and Outpainting), visually confirming our framework's strong and consistent performance across these diverse domains.

\begin{figure*}[ht]
\centering
\includegraphics[width=0.65\linewidth]{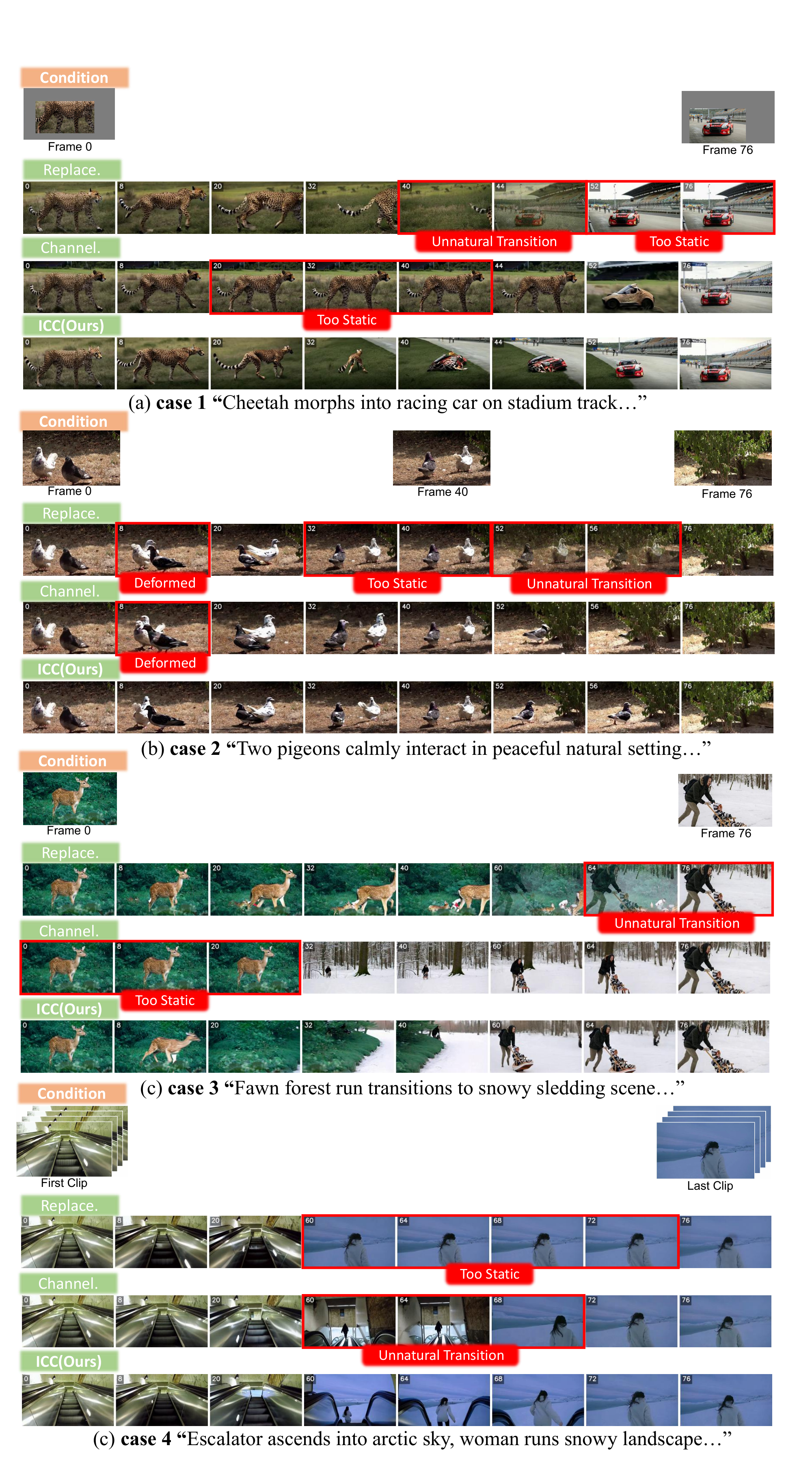}
\caption{\textbf{Comparisons with baseline paradigms.} }
\label{fig:comparision}
\end{figure*}

Finally, Figure~\ref{fig:comparision} provides additional direct comparisons against different paradigms across a diverse set of challenging cases, further highlighting the robustness and superiority of our approach.

\end{document}